\documentclass[runningheads]{llncs}

\usepackage{eccv}

\usepackage{eccvabbrv}

\usepackage{graphicx}
\usepackage{booktabs}
\usepackage{pifont}
\usepackage{multirow}
\usepackage{algorithm}
\usepackage{algpseudocode}

\usepackage[accsupp]{axessibility}  %

\usepackage{hyperref}

\usepackage{orcidlink}

\algnewcommand\algorithmicinput{\textbf{Input:}}
\algnewcommand\algorithmicoutput{\textbf{Output:}}
\algnewcommand\Input{\item[\algorithmicinput]}%
\algnewcommand\Output{\item[\algorithmicoutput]}%

\begin{document}

\title{Source-Free Domain Adaptation for YOLO Object Detection}

\author{Simon Varailhon \and
Masih Aminbeidokhti\orcidlink{0000-0003-2289-3690} \and
Marco Pedersoli\orcidlink{0000-0002-7601-8640} \and \\
Eric Granger\orcidlink{0000-0001-6116-7945}} 

\authorrunning{S.~Varailhon et al.}

\institute{LIVIA, ILLS, Dept. of Systems Engineering, ETS Montreal, Canada \\
\email{\{simon.varailhon.1, masih.aminbeidokhti.1\}@ens.etsmtl.ca\\ \{marco.pedersoli, eric.granger\}@etsmtl.ca}}

\maketitle

\begin{abstract}
Source-free domain adaptation (SFDA) is a challenging problem in object detection, where a pre-trained source model is adapted to a new target domain without using any source domain data for privacy and efficiency reasons. Most state-of-the-art SFDA methods for object detection have been proposed for Faster-RCNN, a detector that is known to have high computational complexity. This paper focuses on domain adaptation techniques for real-world vision systems, particularly for the YOLO family of single-shot detectors known for their fast baselines and practical applications. Our proposed SFDA method -- Source-Free YOLO (SF-YOLO) -- relies on a teacher-student framework in which the student receives images with a learned, target domain-specific augmentation, allowing the model to be trained with only unlabeled target data and without requiring feature alignment. A challenge with self-training using a mean-teacher architecture in the absence of labels is the rapid decline of accuracy due to noisy or drifting pseudo-labels. To address this issue, a teacher-to-student communication mechanism is introduced to help stabilize the training and reduce the reliance on annotated target data for model selection. Despite its simplicity, our approach is competitive with state-of-the-art detectors on several challenging benchmark datasets, even sometimes outperforming methods that use source data for adaptation. Our code is available at \url{https://github.com/vs-cv/sf-yolo}. 

\end{abstract}

\section{Introduction}
\label{sec:intro}

Deep learning models for object detection (OD) have shown impressive performance by leveraging a large amount of labeled data \cite{Ren2015FasterRT,glenn_jocher_2022_7347926,Carion2020EndtoEndOD,Zhang2022DINODW}. Yet, they still perform best when the test distribution is close to the training one. However, in real-world applications, there is typically a shift between data from the source (laboratory) and target (operational) domains due to variations in camera capture conditions (pose, illumination, resolution, weather) that can decrease accuracy   \cite{chen2018domain,Saito2018StrongWeakDA,Oza2021UnsupervisedDA}. Fine-tuning the OD model for the target domain is effective but often too costly due to the need for collecting and annotating data.

Unsupervised domain adaptation (UDA) adapts a model trained on a source domain to perform well on a target domain using unlabeled target data. It mitigated the decline in accuracy caused by domain shift without requiring target data annotation. Traditional UDA methods assume access to labeled source and unlabeled target data, but privacy, confidentiality, and logistical challenges may prevent access to source data. To address this, source-free domain adaptation (SFDA) methods for OD \cite{li2021free, huang2021model, Li_2022_CVPR, vibashan2023instance, chu2023adversarial, Liu2023PeriodicallyET} have emerged, allowing for adaptation without source data, though this makes it more challenging to align source and target distributions.

\begin{table}[!tb]
    \centering
    \caption{Complexity and accuracy of Faster R-CNN and YOLOv5 (large and small) detectors commonly used in UDA studies. mAP accuracy of the \textit{source only} model trained on Cityscapes is shown on Foggy Cityscapes test set, without applying SFDA. The number of frames per second (FPS) is measured on an NVIDIA V100 GPU.}
    \label{tab:comparison}
    \resizebox{0.8\textwidth}{!}{
    \begin{tabular}{
        c
        @{\hspace{4pt}} |
        @{\hspace{4pt}}
        c
        @{\hspace{8pt}}
        c
        @{\hspace{8pt}}
        c
        @{\hspace{8pt}}
        c
    }
    \toprule
    \textbf{Detector} & \textbf{Parameters (M)} & \textbf{FLOPs (G)} & \textbf{FPS} &  \textbf{mAP} \\
    \midrule 
    Faster R-CNN \cite{Ren2015FasterRT,wu2019detectron2} & 34.0  & 822.0 & \ \ 9.8 & 25.2 \\
    YOLOv5l \cite{glenn_jocher_2022_7347926} & 46.5 & 109.1 &  \ 99.0 & 41.8\\
    YOLOv5s \cite{glenn_jocher_2022_7347926}& \ 7.2 & \ 16.5 & 156.3 & 27.4 \\ 
    \bottomrule
    \end{tabular}
    }
\end{table}

The first UDA method for OD  \cite{chen2018domain} was proposed for Faster-RCNN \cite{Ren2015FasterRT}, a two-stage detector. To compare with previous works, most approaches have followed this pioneering work \cite{Saito2018StrongWeakDA,Inoue2018CrossDomainWO,He2019MultiAdversarialFF,Kim2019DiversifyAM,Wang2021AFANAF,Hsu2019ProgressiveDA,chen2021scale,Guan2021UncertaintyAwareUD,Zhuang2020iFANIF,Khodabandeh2019ARL,Li2021CategoryDG,Cai2019ExploringOR} and still rely on Faster-RCNN. With newer detectors gaining popularity, some work \cite{Hsu2020EveryPM,Vidit2021AttentionbasedDA,Zhou2022SSDAYOLOSD} focused on developing UDA methods for single-stage detectors like FCOS \cite{Tian2019FCOSFC} and YOLOv5 \cite{glenn_jocher_2022_7347926}. Yet, newer SFDA methods still use Faster-RCNN \cite{li2021free,huang2021model,Li_2022_CVPR,vibashan2023instance,chu2023adversarial,Liu2023PeriodicallyET}, which may no longer be ideal \cite{Zhou2022SSDAYOLOSD}. Although some SFDA strategies may apply to other detectors, the lack of Region Proposal Network (RPN) in single-stage models, which helps align instance-level features \cite{chen2018domain,Saito2018StrongWeakDA}, can complicate UDA \cite{lee2022rethinking}. Moreover, UDA progress often parallels advancements in OD methods and designs. However, Faster-RCNN is somewhat outdated, computationally intensive, and less suitable for real-time applications.  \Cref{tab:comparison} shows that even without UDA, YOLOv5l is 10 times faster with better performance, while YOLOv5s achieves similar performance but is 15 times faster.

The mean-teacher (MT) framework is a prominent approach for SFDA in OD \cite {vibashan2023instance,zhang2023refined}, using self-distillation and self-training. The student and teacher models start from the same initial model, with the student updated via pseudo-labels from the teacher and the teacher via an exponential moving average (EMA) of past students \cite{Tarvainen2017MeanTA}. The MT framework assumes the teacher model can be continuously improved as training progresses, and the student can gradually approach the teacher's performance.
However, the source-pretrained teacher introduces inherent bias when applied to the target domain due to domain shift \cite{Liu2023PeriodicallyET}.
This bias, combined with student error accumulation often causes instability, degrading both models' performance.
Adjusting EMA hyperparameters for a more gradual and stable knowledge transfer between student and teacher can help stabilize training. Since real-world SFDA applications often have limited test sets, if any, ideal methods should be robust to hyperparameters and require minimal tuning.

In this paper, we explore a different direction than existing SFDA approaches for OD. Inspired by the work of Zhou \etal~\cite{Zhou2022SSDAYOLOSD} on UDA methods with YOLO, we focus on SFDA beyond Faster-RCNN since it is complex and unsuitable for practical scenarios like real-time video surveillance. An SFDA method is proposed for YOLO that does not rely on feature alignment but on a MT framework equipped with a learned target domain-specific data augmentation. To the best of our knowledge, no SFDA approach has been proposed for one-stage detectors such as YOLO, with the majority (if not all) focusing on Faster-RCNN. Additionally, we propose a novel Student Stabilisation Module (SSM) to improve the MT framework's self-training paradigm. As shown in \Cref{fig:architecture}, SSM enables a bidirectional communication channel between the teacher and student models. Unlike previous methods where only the teacher is updated by the student model, our proposed method periodically replaces the student by the moving average of the teacher. This prevents a rapid decrease in the lower bound of the student model, ensuring the robustness of our entire pipeline. Our method is highly robust to the choice of hyperparameter values, and therefore suitable for real-world applications of SFDA. We conduct several ablations to empirically validate the effectiveness of our proposed teacher-student framework with SSM, highlighting the critical role of detector selection and training stability in SFDA.

\noindent \textbf{Our main contributions are summarized as follows.}\\
\noindent \textbf{(1)} We introduce Source-Free YOLO (SF-YOLO), the first SFDA method specialized for one-stage YOLO detectors, establishing a baseline for future research targeting practical real-time applications. Our proposed SF-YOLO approach utilizes a teacher-student framework with a learned target domain-specific augmentation module that allows for training using only unlabeled target domain data and without requiring feature alignment.\\
\noindent \textbf{(2)} A Student Stabilisation Module (SSM) is proposed to mitigate the training instability and related accuracy degradation caused by the absence of labeled data when using the mean teacher paradigm. It provides a new communication channel from teacher to student that enhances training stability, and thereby reduces the reliance on annotated target data for model selection.\\
\noindent \textbf{(3)} Our extensive experiments show the effectiveness of our proposed SFDA method for YOLO on several challenging domain adaptation benchmarks with Cityscapes, Foggy Cityscapes, Sim10k, and KITTI datasets. We evaluate our method using two variants of the YOLOv5 model -- a more accurate but larger model and a smaller model for efficiency. Our approach outperforms state-of-the-art detectors yet requires low computational resources and can achieve competitive performance against UDA methods that require source data for adaptation.

\section{Related Work}

\noindent \textbf{(a) Unsupervised Domain Adaptation.} 
UDA aims to adapt a source model to perform well in a target (operational) domain using labeled source data and unlabeled target data. UDA methods for OD cover both classification and localization tasks and are typically classified as adversarial feature learning, domain translation, or self-training \cite{Oza2021UnsupervisedDA}. Adversarial methods use domain discriminators and adversarial training \cite{Ganin2015DomainAdversarialTO} to confuse the model between the source and target domains, resulting in a domain-invariant representation. Chen \etal~\cite{chen2018domain} first introduced image and instance-level adversarial feature alignment. Saito \etal~\cite{Saito2018StrongWeakDA} refined this with strong local and weak global feature alignment, both employing adversarial techniques.
Building on these ideas, some works \cite{Zhuang2020iFANIF, Guan2021UncertaintyAwareUD} focused on instance-level alignment, emphasizing regions containing objects of interest.

In contrast to adversarial image or instance-level alignment methods \cite{Belal_2024_WACV,chen2018domain,Saito2018StrongWeakDA,Inoue2018CrossDomainWO,He2019MultiAdversarialFF,Hsu2019ProgressiveDA}, Hsu \etal~\cite{Hsu2020EveryPM} proposed a center-aware feature alignment method, using the FCOS detector. Alternatively, domain translation methods  \cite{Saito2018StrongWeakDA,Kim2019DiversifyAM,Tasar2020StandardGANMD,deng2021unbiased,Wang2021AFANAF,Zhou2022SSDAYOLOSD} typically use image-to-image translation methods like CycleGAN \cite{Zhu2017UnpairedIT}, AdaiN \cite{Huang2017ArbitraryST}, FDA \cite{Yang2020FDAFD}, or CUT \cite{Park2020ContrastiveLF} to learn domain-invariant representation by generating source like target images or target like source images. Self-training methods often use a pseudo-labeling strategy \cite{Khodabandeh2019ARL,Li2021CategoryDG} for supervised fine-tuning of the models in the target domain, but incorrect pseudo-labels can degrade performance. Therefore Kim \etal~\cite{Kim2019SelfTrainingAA} proposes to use a weak self-training approach to mitigate this, while Khodabandeh \etal~\cite{Khodabandeh2019ARL} refine noisy annotations with an additional classifier. To prevent student-teacher collapse, Lin \etal \cite{lin2023run} add an extra domain alignment loss to distinguish between the student and teacher models. However, all of the mentioned methods cannot meet the growing demand for data privacy protection. SFDA has emerged as a new branch of UDA.

\noindent \textbf{(b) Source-Free Domain Adaptation.} 
SFDA adapts a source model pre-trained on labeled source data using only unlabeled target data, addressing privacy concerns by transmitting the model instead of extensive source-domain data. SFDA methods for OD mostly use MT and a pseudo-labeling strategy \cite {vibashan2023instance, Li_2022_CVPR, Liu2023PeriodicallyET, Bochkovskiy2020YOLOv4OS}.
A key challenge of fine-tuning methods for SFDA is their reliance on noisy pseudo-labels. Since inaccurate pseudo-labels can degrade model accuracy, Li \etal~\cite{li2021free} first addressed this issue by determining a threshold for eliminating unreliable pseudo-labels using an adaptive entropy minimization approach. Zhang \etal~\cite{zhang2023refined} refined this by adapting thresholds per class and improving the localization quality of pseudo-labels. A$^{2}$SFOD \cite{chu2023adversarial} tackles noisy labels by partitioning the target domain based on detection variance. Huang \etal~\cite{huang2021model} use self-supervised learning for learning feature representations through historical models and contrastive learning while Vibashan \etal \cite{vibashan2023instance} employ instance relation graph networks and a contrastive loss to enhance target representations.

Li \etal~\cite{Li_2022_CVPR}, argued that existing SFDA methods do not fully utilize target domain data, limiting their effectiveness. To address this, they proposed LODS, which reduces the model's focus on domain style by enhancing the style of target domain images and using the style difference between the original and enhanced images as a self-supervised signal for adaptation. Most approaches rely on a MT architecture, which is prone to instability,  limiting the performance they can reach. Closely related to our method, Liu \etal~\cite{Liu2023PeriodicallyET} introduced PETS, which exchanges teacher and student models each epoch while maintaining an additional EMA teacher. However, PETS requires an extra model and relies on weighted box fusion which increases the complexity of the pipeline. Moreover, as shown in \Cref{sec:exp}, with our implementation on YOLO (as we could not find the public one), it does not substantially improve the SFDA accuracy with the YOLO architecture. Overall, while the mentioned methods use Faster R-CNN, we introduce a new SFDA approach with YOLO and our SSM, providing more stable training and improved performance for real-world applications.

In this paper, we focus on the MT architecture and use a learned data augmentation as proposed by Li \etal \cite{Li_2022_CVPR}, which leverages target domain data, rather than the typical strong and weak augmentation pairs \cite{vibashan2023instance,Liu2023PeriodicallyET,lin2023run,zhang2023refined}. Furthermore, we address the training instability \cite{Liu2023PeriodicallyET} of the MT in the absence of labeled data by introducing a teacher-to-student communication mechanism to stabilize training, which is crucial in SFDA due to the lack of validation data. We show through our ablations that our proposed approach significantly improves the accuracy of the MT framework while remaining simple and straightforward. 

\noindent \textbf{(c) UDA for YOLO.}
Faster-RCNN is a two-stage OD model with separate branches for classification and localization. This is convenient for UDA, in particular for feature alignment \cite{chen2018domain,Saito2018StrongWeakDA}. However, due to its high computational complexity, it's rarely used in industrial applications requiring real-time processing \cite{Li2023ADA}. To create UDA methods better suited for real-time tasks, some work \cite{Zhang2021DomainAY,Vidit2021AttentionbasedDA,Hnewa2021MultiscaleDA,Zhou2022SSDAYOLOSD,wei2023yolo} focused on YOLO detectors \cite{Redmon2018YOLOv3AI,Bochkovskiy2020YOLOv4OS,glenn_jocher_2022_7347926}.
SFDA methods for YOLO can be categorized into feature alignment and data-based methods. 

Among feature alignment methods, MS-DAYOLO \cite{Hnewa2021MultiscaleDA} uses multi-scale image-level adaptation but does not perform local alignment, which is effective \cite{chen2018domain}. DA-YOLO \cite{Zhang2021DomainAY} addresses this by leveraging image-level and instance-level feature alignment, replacing the RPN with a detection layer on YOLO's three-scale feature maps. 
Vidit \etal~\cite{Vidit2021AttentionbasedDA} 
use an attention mechanism to identify the foreground regions where adaptation should be performed, replacing the RPN. CAST-YOLO \cite{liu2023cast} also employs attention, but for aligning source and target features within a MT architecture. SSDA-YOLO \cite{Zhou2022SSDAYOLOSD} combines image-to-image translation and feature alignment using a MT framework, leveraging CUT \cite{Park2020ContrastiveLF} to generate source-like target images and vice-versa. Pseudo-labels from the teacher model enable instance-level alignment, and a consistency loss aligns predictions across domains. However, in our SFDA paradigm, methods relying on feature alignment between the source and target domains, such as adversarial learning \cite{Hnewa2021MultiscaleDA,Zhang2021DomainAY,wei2023yolo}, are not directly applicable due to the lack of source data. 

Data-based methods, focus more on data augmentation. ConfMix \cite{Mattolin2022ConfMixUD} introduces a sample mixing strategy, combining one source image with a crop of the target image with the highest detection confidence. They also use a pseudo-labeling strategy with a variable threshold during training. SimROD \cite{Ramamonjison2021SimRODAS} introduces DomainMix, which mixes labeled source images with unlabeled target images and their pseudo-labels, generated by a large teacher model. To the best of our knowledge, no YOLO-based method has been proposed for the SFDA paradigm. Therefore, we propose a method that also employs data augmentation for adaptation rather than feature alignment and does not rely on a specific architecture, making it more versatile.

\section{Proposed Method}

Let $\mathcal{D}_s = (\mathcal{X}_s, \mathcal{Y}_s)$ represent the labeled data in the source domain where $\mathcal{X}_s = \{x_{s}^i \}_{i=1}^{N_s}$ denotes the
image set of the source domain and $\mathcal{Y}_s = \{y_{s}^i \}_{i=1}^{N_s}$ is the corresponding label set containing object locations and category assignments for each image. $N_s$ denotes the total number of source images. The target domain $D_t = (\mathcal{X}_t$) is unlabeled and $\mathcal{X}_t = \{x_{t}^i \}_{i=1}^{N_t}$ denotes the $N_t$ images of this domain. 

In SFDA, a source pre-trained model denoted as $h_s: \mathcal{X}_s \rightarrow \mathcal{Y}_s$ is initially available to perform adaptation on the unlabeled target domain. However, given the domain shift between the source and target domains, the mapping $h_s$ diminishes performance when directly applied to the target domain. Consequently, the primary objective of SFDA is to learn a new mapping $h_t: \mathcal{X}_t \rightarrow \mathcal{Y}_t$ by adapting the pre-trained source model 
$h_s$ using only the unlabeled target data $\mathcal{X}_t$.

Using the mean-teacher paradigm, both teacher, defined as $h_\theta$, and student defined as $h_\phi$ models are instantiated from the source model which is a pre-trained YOLOv5 architecture. Therefore the training loss can be expressed as: 
\begin{equation}
    \mathcal{L}_{\text{det}} = \lambda_{\text{b}} \mathcal{L}_{\text{box}} + \lambda_{\text{c}}
    \mathcal{L}_{\text{cls}} + \lambda_{\text{d}} \mathcal{L}_{\text{obj}} \ , 
    \label{eq:yolo_loss}
\end{equation}
where $\mathcal{L}_{\text{box}}$ and $\mathcal{L}_{\text{cls}}$ respectively represent the classification and bounding box regression losses, while $\mathcal{L}_{\text{obj}}$ corresponds to the objectiveness loss, which is related to the confidence of object presence. The $\lambda_{\text{b}}$, $\lambda_{\text{c}}$, $\lambda_{\text{d}}$ terms are weighting hyperparameters that control the relative importance of each loss component in the overall detection loss.

\begin{figure}[!t]
  \centering
  \includegraphics[width=\linewidth]{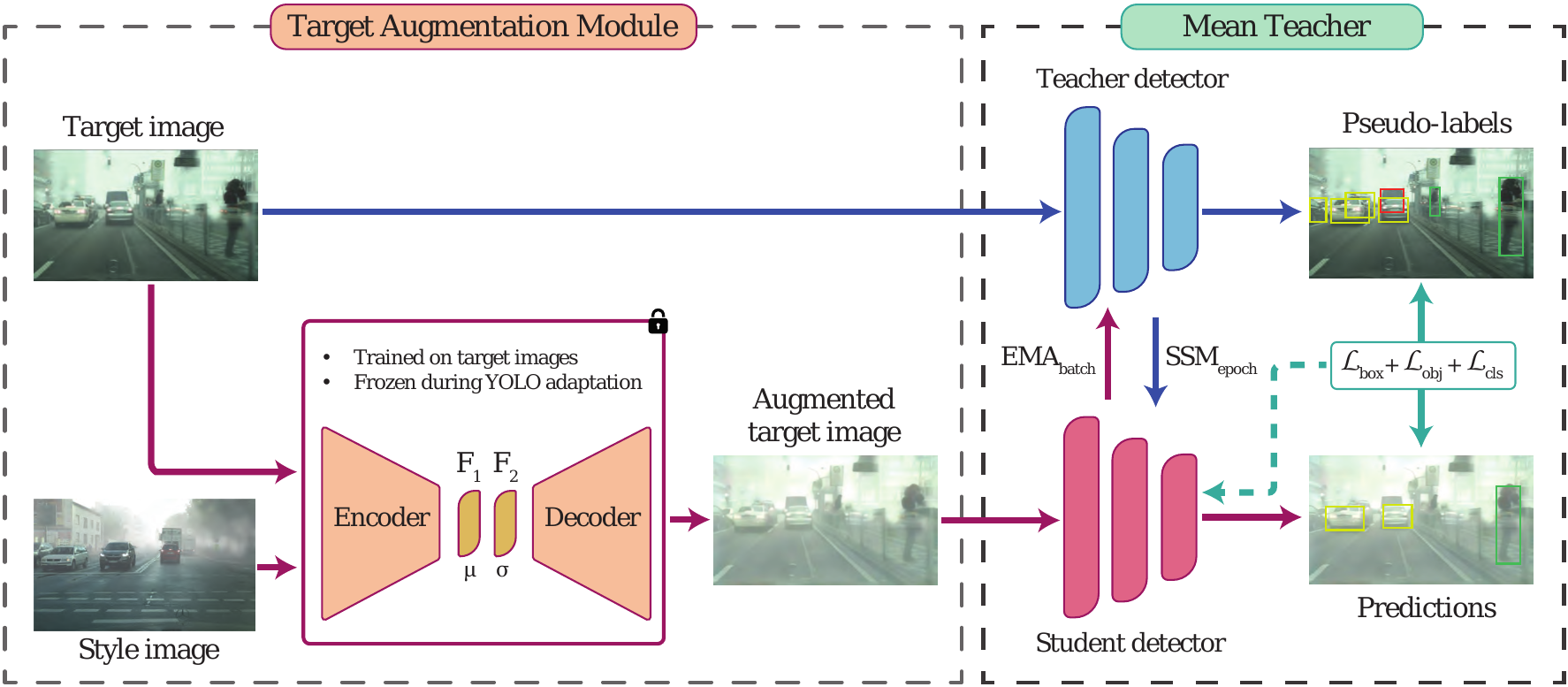} 
  \caption{The proposed SF-YOLO training architecture. First, TAM is trained using all the target images from the training set. These learned augmented images are then used as data augmentation for the student while the teacher receives unmodified target images. The student detector learns through backpropagation, as per \cref{eq:yolo_loss}, and then updates the teacher detector using EMA for each batch. Finally, at a lower frequency, once every epoch, the teacher updates the student with SSM to stabilize the training.}
  \label{fig:architecture} 
\end{figure}

Our framework works in two steps. First, a Target Augmentation Module is trained to learn a domain-specific data augmentation for the target domain. Then, we use the augmented images to train a student model with the mean teacher paradigm. The student model $h_{\phi}$ gradually distills its acquired knowledge to the teacher model $h_{\theta}$ which learns through exponential moving average (EMA). We also introduce the Student Stabilisation Module (SSM), which effectively stabilizes the training process, leading to improved overall performance. The framework does not increase the basic detector's complexity during inference. This is an important factor for many real-time applications. The overall SF-YOLO training architecture is illustrated in \Cref{fig:architecture} and detailed in \cref{alg:cap}. The rest of this section describes the main components of our method.

\noindent \textbf{(a) Target Augmentation Module.}
In typical self-training with the teacher-student framework, the teacher and student models take distinct views of images as input to maximize mutual information in a non-trivial manner. Specifically, The teacher model receives weakly-augmented images as input, while the student model is fed with strongly-augmented images. Here, instead of generating random strong-weak augmentation pairs, we use a network coined "Target Augmentation Module" (TAM) to learn the proper augmentation. This augmentation module reminisces the style enhancement module from the LODS method \cite{Li_2022_CVPR} with the main difference being motivation. In our case, we use the augmentation module to enrich the target domain, whereas in LODS, they use a style enhancement module to learn to overlook the target domain style. Here we briefly overview the main architecture of the TAM. 

Suppose we have an image from the target domain, denoted by $x$, and a style image $y$, which represents either the average of all target domain images if the backgrounds are similar, or a random target image. The TAM transforms the input image based on the style image statistics following the style transfer formulation \cite{Huang2017ArbitraryST}:
\begin{equation}
e_x^{y} = \sigma(e_y) \frac{e_x - F_1(\mu(e_x), \mu(e_y))}{F_2(\sigma(e_x), \sigma(e_y))} + \mu(e_y),
\end{equation}
where $e_x$ and $e_y$ are the VGG-16 encoded features for $x$ and $y$ respectively. $\mu(e_x)$ and $\sigma(e_x)$ denote the channel-wise mean and variance of $e_x$, and similarly for $e_y$. $F_1$ and $F_2$ are neural networks that combine the means and variances of $x$ and $y$. We follow the same architecture as Li \etal \cite{Li_2022_CVPR} for $F_1$ and $F_2$. The TAM is trained by minimizing the style consistency and reconstruction losses.

\noindent \textbf{(b) Consistency Learning With Teacher Knowledge.}
For each data point $x$ from the target domain $D_t$ we obtain the classification score $h_{\theta}^{\text{cls}}(x)$, bounding box regression $h_{\theta}^{\text{box}}(x)$ and confidence $h_{\theta}^{\text{obj}}(x)$ from the teacher model, and use it as the pseudo-label to train the student model via backpropagation. However, directly learning from $h_{\theta}(x)$ may lead to overfitting to the teacher model. To address this issue, we introduce a consistency loss that consists of three steps. First, we filter out the low-confidence predictions of the teacher models by setting a classification confidence threshold $\delta$. This can prevent the subsequent process from suffering from the interference of noisy labels. Next, we augment $x$ using the TAM and train the student model to be consistent with the hard-label $p$ for the augmented samples. Third, the student gradually distills its acquired knowledge to the teacher model which learns through exponential moving average (EMA):
\begin{equation}
\theta = \alpha \theta + (1-\alpha)\phi
\label{eq:mean_teacher}
\end{equation}
where $\alpha$ is the EMA momentum parameter.

\noindent \textbf{(c) Student Stabilisation Module.}
Note that the teacher never directly observes the augmented images, which can often be unrealistic. Instead, it only deals with the real target images. So far our method is similar to the Mean Teacher and Noisy Student paradigm \cite{Tarvainen2017MeanTA}. However, since the student model learns faster, it is more susceptible to making errors. After updating the teacher, those mistakes are reflected in pseudo-labels generation, causing noisier pseudo-labels which can degrade performance for the rest of the training. To mitigate the impact of these errors, the Student Stabilisation Module (SSM) is introduced.
\begin{figure}[!t]
  \centering
  \subfloat[$\eta$ = 0.01]{\includegraphics[width=0.3\linewidth]{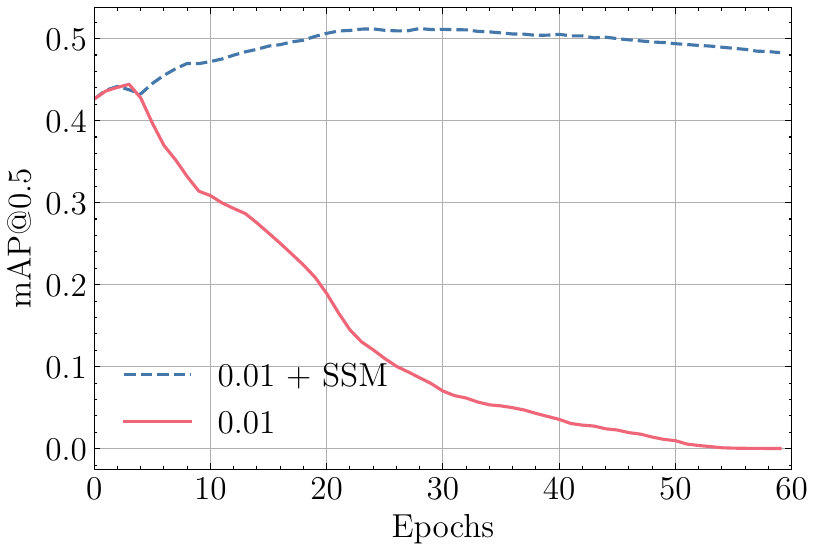} \label{fig:lr0}} \quad
  \subfloat[$\eta$ = 0.008]{\includegraphics[width=0.3\linewidth]{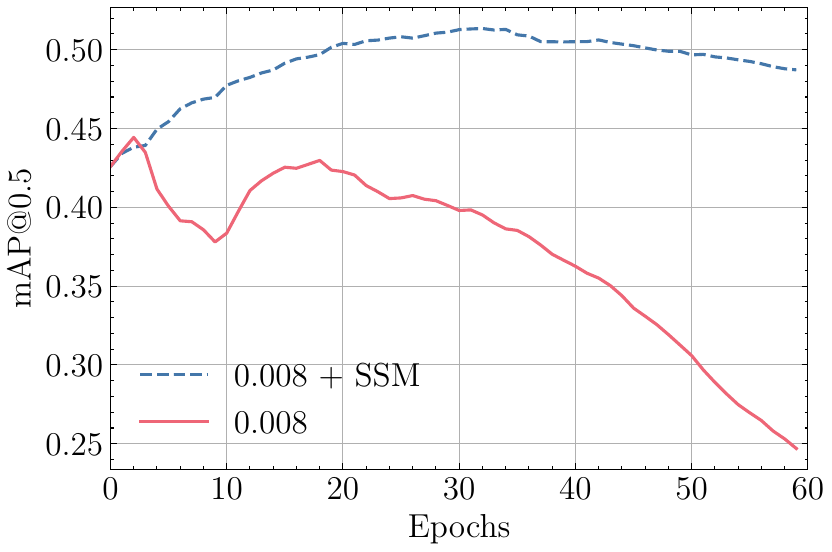} \label{fig:lr1}} \quad
  \subfloat[$\eta$ = 0.006]{\includegraphics[width=0.3\linewidth]{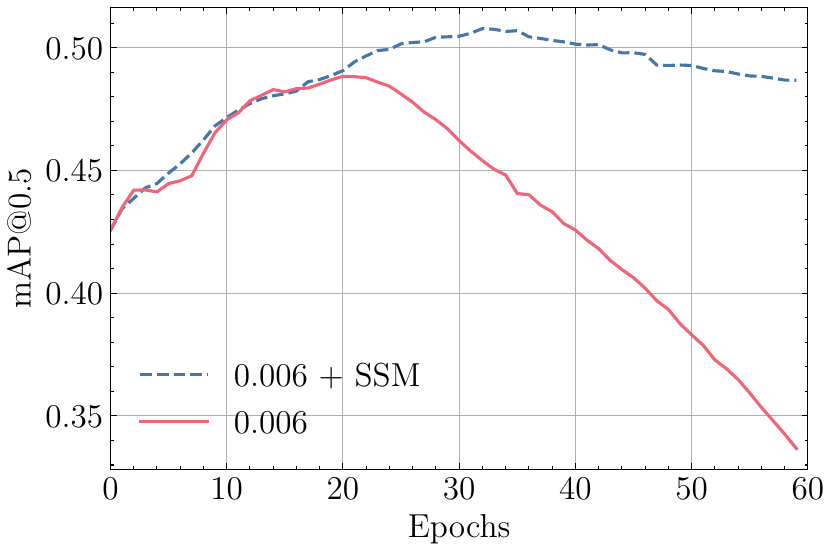} \label{fig:lr2}} \\
  \caption{The training curves of the target augmented mean-teacher with and without SSM using YOLOv5l on the C2F scenario across different learning rates. SSM prevents MT from quick deterioration and reaches better final performance.}
  \label{fig:sec-3-ssm_yolov5l}
\end{figure}

As our initial experiments using the target augmented mean-teacher framework in \Cref{fig:sec-3-ssm_yolov5l} shows, although EMA and learned augmentation help to improve the performance of the student model, the result quickly deteriorated to the extent that at epoch 60 the performance of student model is almost like a random detector. We hypothesize that since the student model is updated rapidly using SGD after every batch, it is more likely to make errors. Although an EMA-updated teacher can initially prevent the student from overfitting, it may not be sufficient to stabilize the training. As a remedy once at the end of each epoch, we also update the student weights using the EMA of the teacher model using the following equation:    
\begin{equation}
	\phi = \gamma \phi + (1-\gamma)\theta
	\label{eq:rev_mean_teacher}
\end{equation}
where $\gamma$ is the SSM momentum parameter. This approach confines the student model close to the teacher model, maintaining the quality of pseudo-labels and preventing too much deviation. 

\noindent \textbf{(d) Training and Inference.}
The overall SF-YOLO learning process is summarized in \cref{fig:architecture} and \Cref{alg:cap}. The adapted teacher model is used for inference, having learned a more robust representation of the target domain than the student which has not seen any unmodified target domain images. The final model maintains the original YOLO architecture's speed and compatibility benefits.

\begin{algorithm}[!t]
	\caption{The SF-YOLO method}\label{alg:cap}
	\renewcommand{\algorithmicrequire}{\textbf{Input:}}
	\renewcommand{\algorithmicensure}{\textbf{Output:}}
	\begin{algorithmic}
	\Require target domain $D_t$, teacher model $h_\theta$, student model $h_\phi$, Target Augmentation Module TAM, EMA momentum $\alpha$, SSM momentum $\gamma$, confidence threshold $\delta$, learning rate $\eta$, total number of epochs $E$.
	\Ensure teacher model $h_\theta$.
	\For {epoch $t=0$ to $E-1$}
	\For {every minibatch $x$ in $D_t$}
	\State $p = h_{\theta}^{\text{obj}}(x) \geq \delta$;
	\State $\hat{x} = \text{TAM}(x)$;
	\State $\phi \leftarrow \phi - \eta \nabla_{\phi}\mathcal{L}_{\text{det}}(\hat{x}, p, \phi)$; \Comment{Update student according to the \cref{eq:yolo_loss}}
	\State $\theta \leftarrow \alpha \theta + (1-\alpha)\phi$; \Comment{\cref{eq:mean_teacher}}
	\EndFor
	\State $\phi \leftarrow \gamma \phi + (1-\gamma)\theta$; \Comment{\cref{eq:rev_mean_teacher}}
	\EndFor
\end{algorithmic}
\end{algorithm}

\section{Results and Discussion} \label{sec:exp}

\subsection{Experimental Methodology}

\noindent \textbf{(a) Datasets.}
The following four datasets are used: Cityscapes  \cite{Cordts2016TheCD}, Foggy Cityscapes \cite{Sakaridis2017SemanticFS}, KITTI \cite{geiger2012we} and  Sim10k \cite{johnson2016driving}. For more information about the classes considered on each dataset, the domains, the number of images and instances present and other information please refer to the Suppl. Materials.

\noindent \textbf{(b) Adaptation scenarios.} 
Following prior works \cite{chen2018domain, Hsu2020EveryPM,li2021free,Zhang2021DomainAY,Mattolin2022ConfMixUD,vibashan2023instance,chu2023adversarial,wei2023yolo}, we conduct experiments on three different benchmark scenarios: \textit{Cityscapes $\rightarrow$ Foggy Cityscapes} (C2F) for normal weather to adverse weather adaptation, \textit{KITTI $\rightarrow$ Cityscapes} (K2C) to address cross-camera adaptation and \textit{Sim10K $\rightarrow$ Cityscapes} (S2C) to evaluate synthetic-to-real adaptation. As in Sim10k dataset only cars are annotated, we follow the literature \cite{chen2018domain,deng2021unbiased,li2021free,Zhang2021DomainAY,Li_2022_CVPR,vibashan2023instance,chu2023adversarial,wei2023yolo,Li2022CrossDomainOD} and only consider the car AP for S2C and K2C scenarios, while for C2F scenario, we use the complete 8 classes.

\noindent \textbf{(c) Implementation details.} We prioritize a realistic setting, using YOLOv5, a widely adopted and efficient one-stage detector. Additionally, as we are first to address the SFDA paradigm for YOLO, we also conduct a preliminary comparison with baseline UDA approaches \cite{Zhou2022SSDAYOLOSD,wei2023yolo,Mattolin2022ConfMixUD,Li2022CrossDomainOD,liu2023cast} that also use YOLOv5. First, we train the TAM with the same parameters as proposed by the authors of LODS \cite{Li_2022_CVPR}, using Adam \cite{Kingma2014AdamAM} optimizer, a learning rate of 0.0001 and a frozen pretrained VGG16 \cite{Simonyan2014VeryDC} encoder. This module is frozen and used to transform the target images as a data augmentation for the student detection model. 

For source model training we use the default YOLOv5l settings \cite{glenn_jocher_2022_7347926} and train for up to 200 epochs. For our SF-YOLO adaptation, we set the batch size to 16 and trained for 60 epochs. Following Zhou \etal~\cite{Zhou2022SSDAYOLOSD} images are resized to 960×960 and the pseudo-label generation NMS IoU threshold is set to 0.3. We use a confidence threshold $\delta = 0.4$ for the teacher pseudo-labels. The $\alpha$ for the $\text{EMA}_{\phi \rightarrow \theta}$ is set to 0.999 and $\gamma$ used by our SSM module is set to 0.5. Other hyperparameters are the default ones used by YOLOv5. This includes the learning rate which is set to 0.01, using SGD with momentum optimizer. We report the mean Average Precision (mAP) with an IoU threshold set to 0.5, using the validation set of the target domain dataset.

\subsection{Comparison with State-of-the-Art Methods}

Results in \Cref{tab:foggy,tab:combined} show the performance of state-of-the-art SFDA methods based on Faster-RCNN. However, since we are the first to address the SFDA paradigm for YOLO and aim to establish a baseline for future research, SF-YOLO is compared with classical UDA methods based on YOLOv5. Note that these UDA methods access both labeled source data and unlabeled target data, while our method only uses unlabeled target data. Despite this more challenging setup, we achieve competitive performance with UDA methods, suggesting that the source data may not be fully exploited. We also compare with PETS \cite{Liu2023PeriodicallyET}, an SFDA method originally on based on Faster RCNN that we re-implemented (as we could not find the public code). We indicate the use of source data during adaptation in the \textit{Source-free} column. \textit{Source Only} and \textit{Oracle} (upper bound) are respectively trained with labeled source and target data. 

\begin{figure}
    \centering
        \begin{minipage}{.328\textwidth}
            \begin{subfigure}{\textwidth}
            \centering
            \includegraphics[width=\textwidth]{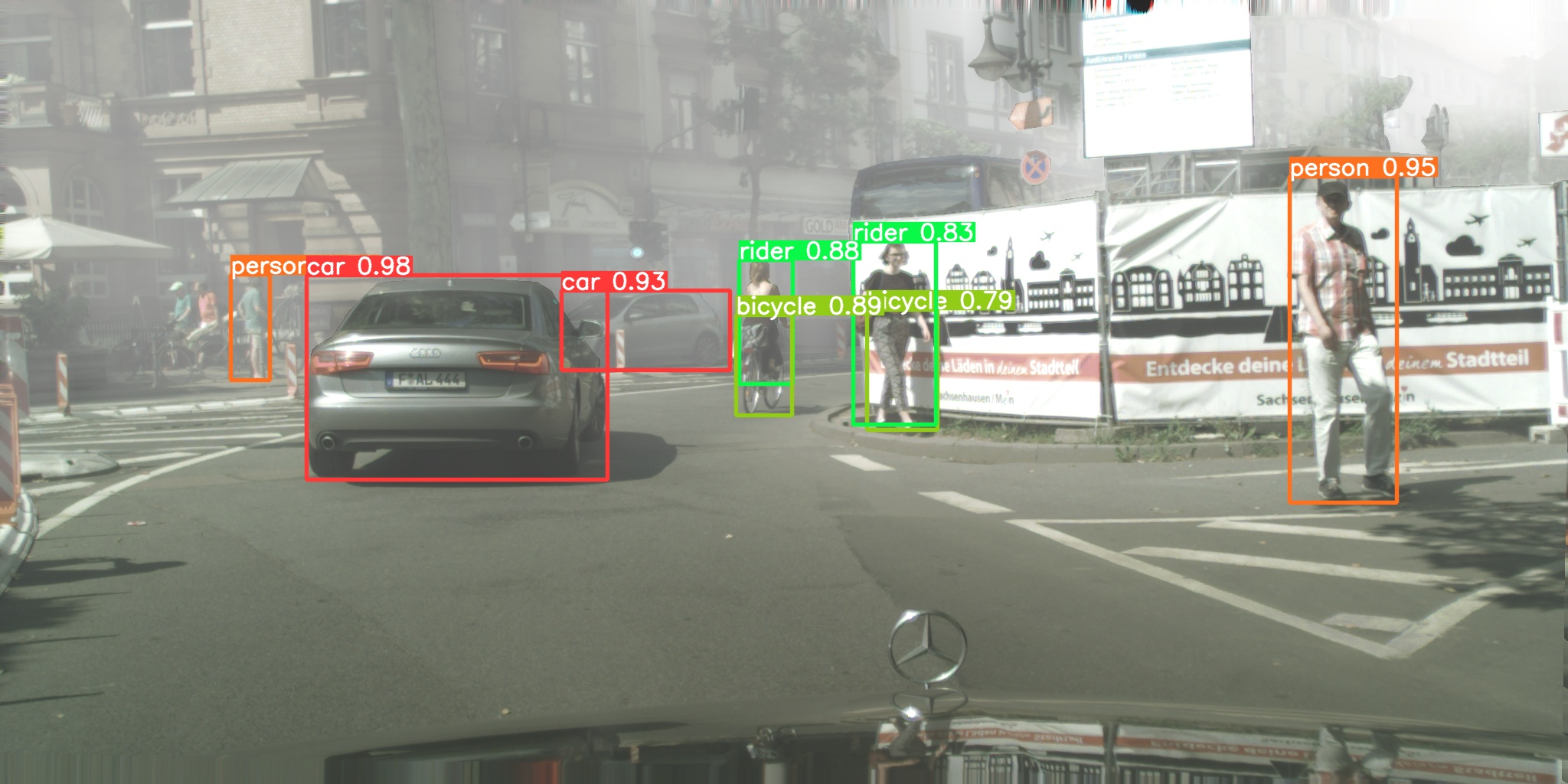}
            \end{subfigure}\\
            \begin{subfigure}{\textwidth}
            \centering
            \includegraphics[width=\textwidth]{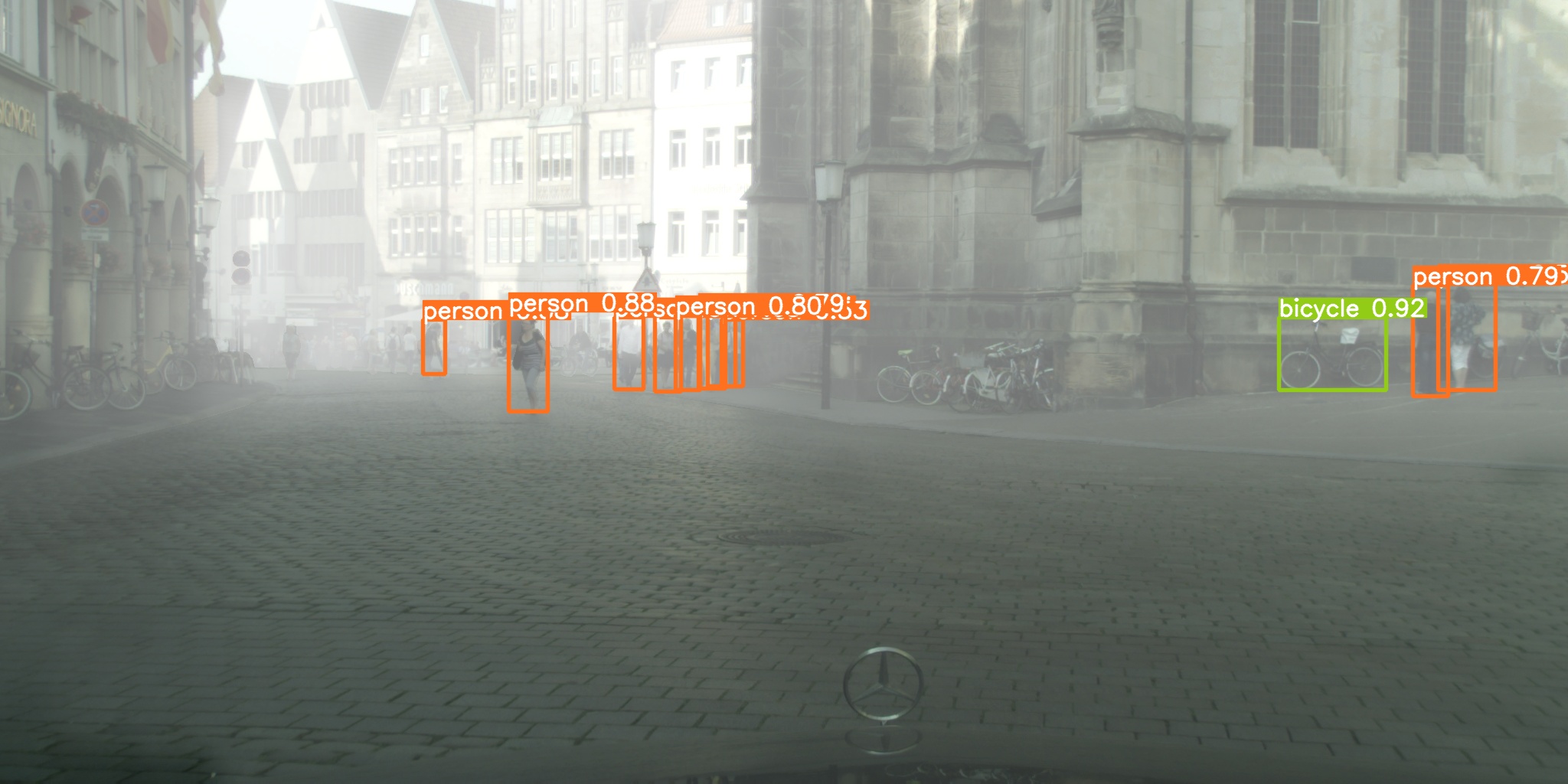}
            \caption{Source-only} \label{fig:source}
            \end{subfigure}%
        \end{minipage}
        \hfill
        \begin{minipage}{.328\textwidth}
            \begin{subfigure}{\textwidth}
            \centering
            \includegraphics[width=\textwidth]{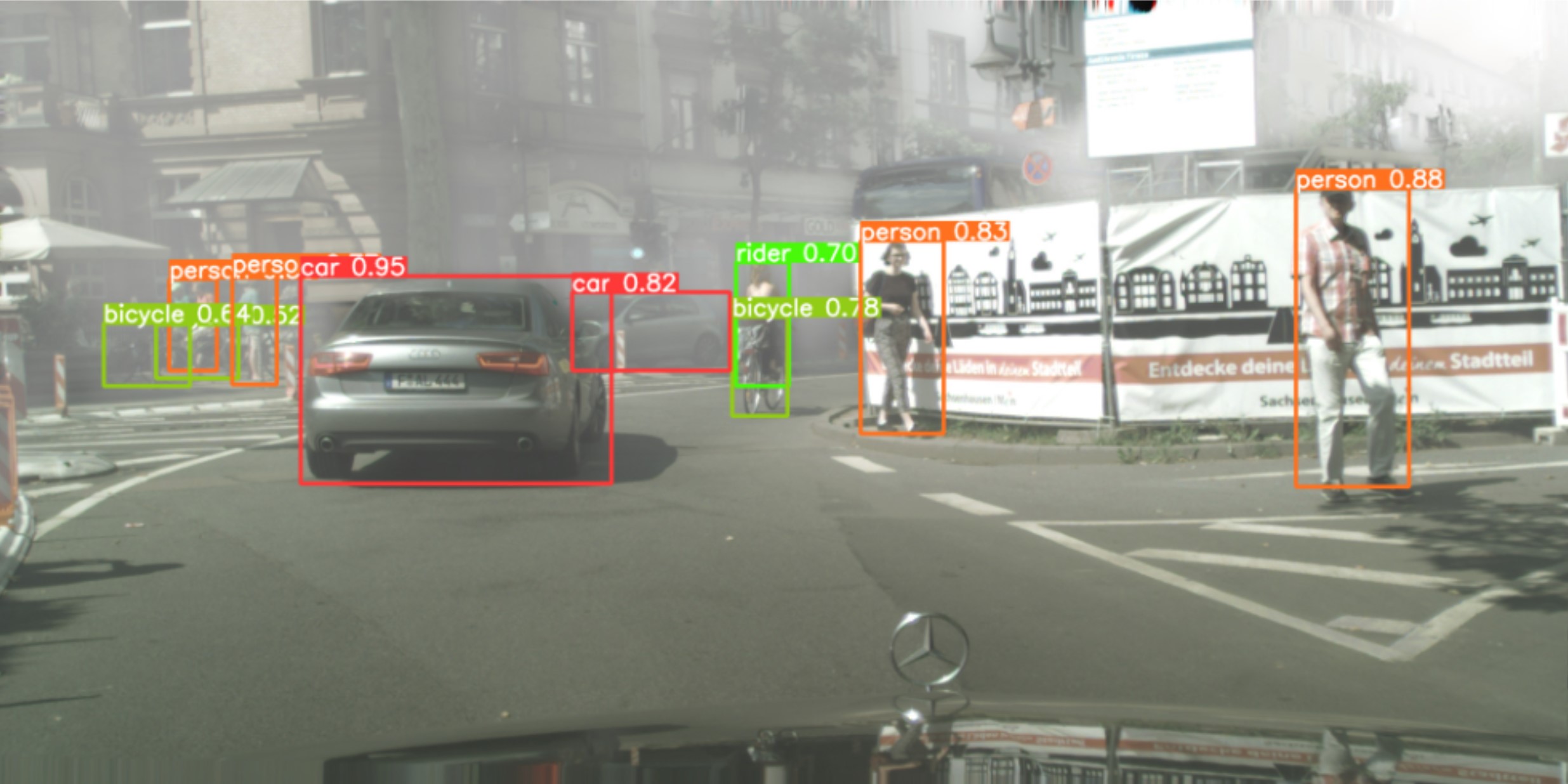} 
            \end{subfigure}\\
            \begin{subfigure}{\textwidth}
            \centering
            \includegraphics[width=\textwidth]{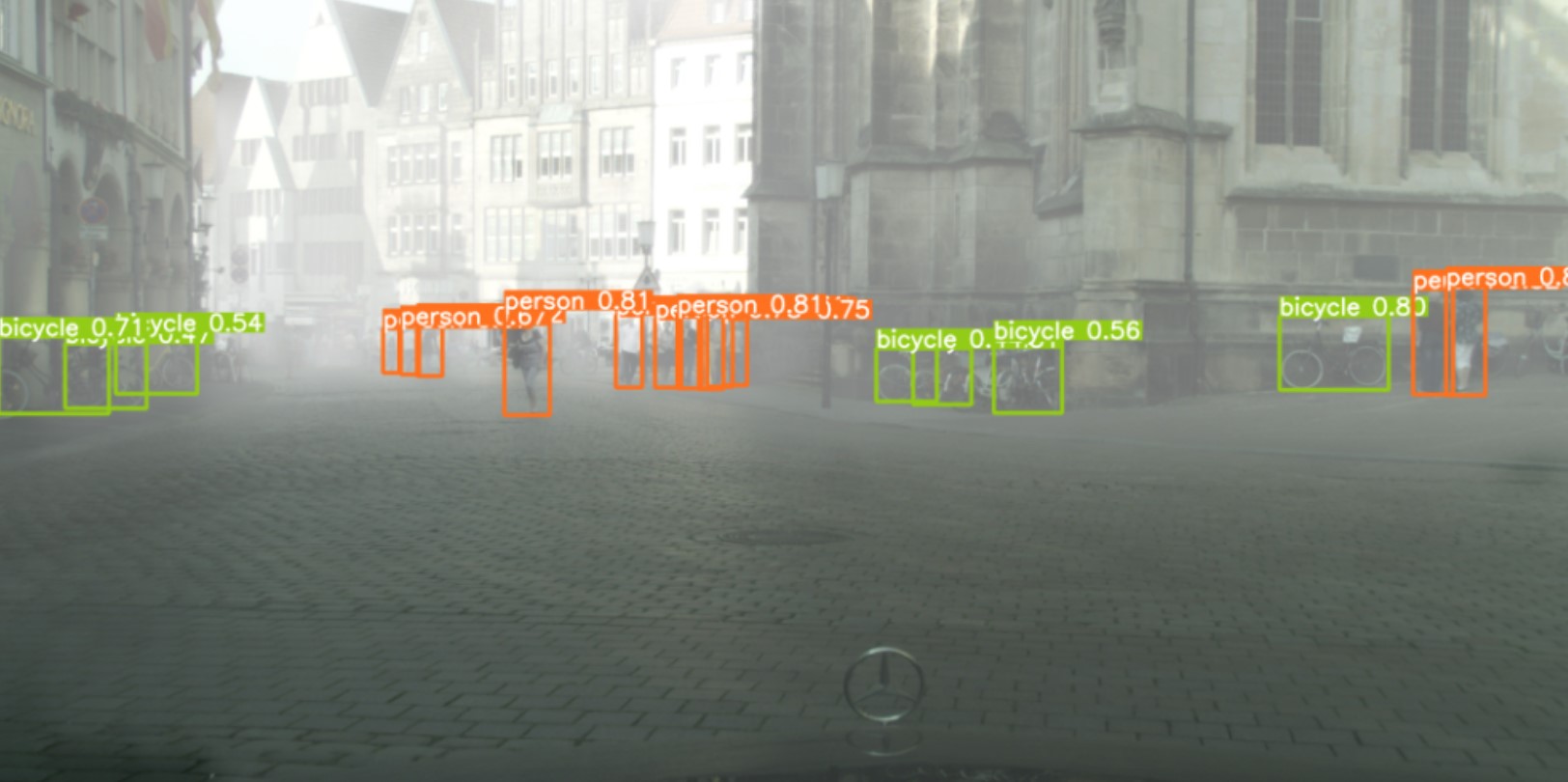}
            \caption{CAST-YOLO \cite{liu2023cast}} 
            \label{fig:cast-yolo}
            \end{subfigure}%
        \end{minipage}
        \hfill
        \begin{minipage}{.328\textwidth}
            \begin{subfigure}{\textwidth}
            \centering
            \includegraphics[width=\textwidth]{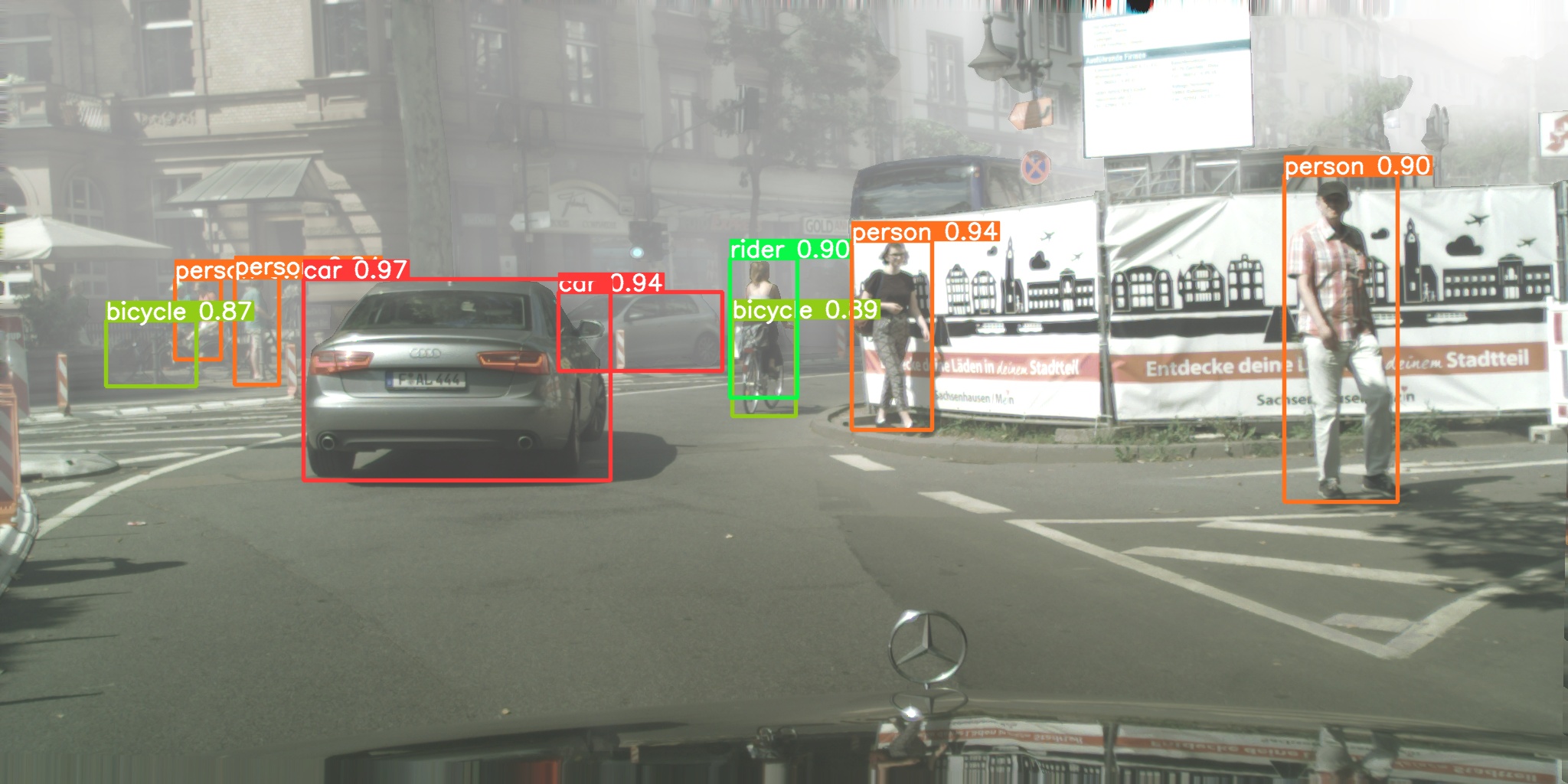}
            \end{subfigure}\\
            \begin{subfigure}{\textwidth}
            \centering
            \includegraphics[width=\textwidth]{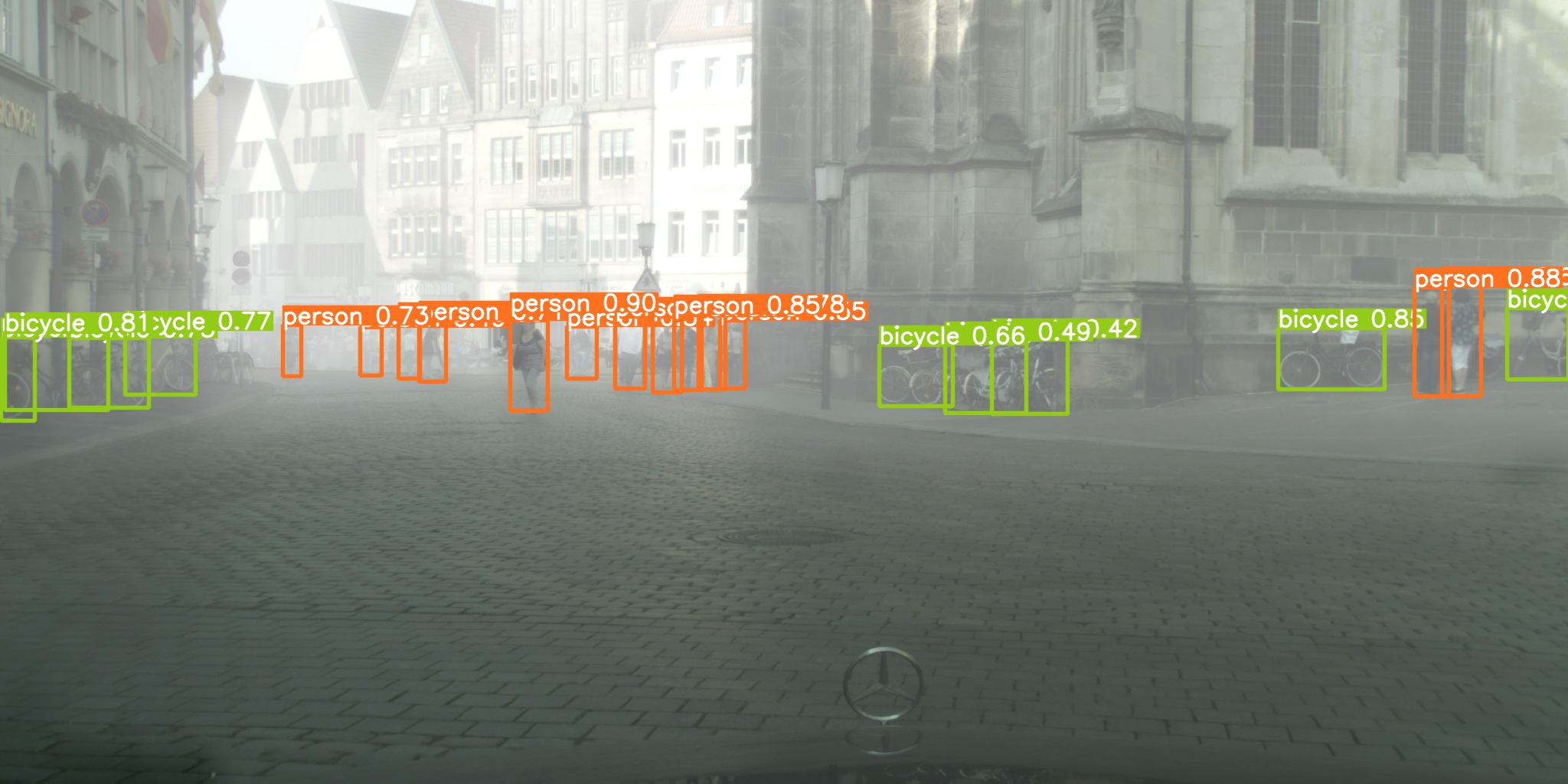}
            \caption{SF-YOLO (Ours)}   \label{fig:ours}        \end{subfigure}%
            
        \end{minipage}%
        \caption{Examples target domain detections for the C2F scenario. Each color represents a class. Our approach qualitatively exhibits comparable performance to CAST-YOLO \cite{liu2023cast} (see \cref{tab:foggy}), yet we did not utilize labeled source data during the adaptation phase.}
        \label{fig:viz}
\end{figure}

\Cref{fig:viz} displays visual detection results for a \textit{source only} model, CAST-YOLO \cite{liu2023cast} and our method. Our SF-YOLO method adapts better to the specific domain, detecting small and blurred objects. 

\begin{table}
	\centering
    \caption{Comparison of our SF-YOLO with state-of-the-art UDA and SFDA methods on the C2F scenario. We report AP50 for each class and mAP. The use of source data during adaptation is denoted by the \textit{Source-free} column. Bold and underlined results represent the best and second-best. We use a default SSM momentum $\gamma$ across all scenarios and also include SF-YOLO$^\dag$ with an optimized $\gamma$. For PETS, we used our implementation on YOLO as we could not find the public one.}
	\label{tab:foggy}
	\resizebox{\textwidth}{!}{%
	\begin{tabular}{
		p{0.3\textwidth}
		c
		c
        @{\hspace{4pt}} | @{\hspace{4pt}} 
		p{0.090\textwidth}
		p{0.090\textwidth}
		p{0.090\textwidth}
		p{0.090\textwidth}
		p{0.090\textwidth}
		p{0.090\textwidth}
		p{0.090\textwidth}
		p{0.090\textwidth}
		p{0.070\textwidth}
		c
	}
	\toprule
	\textbf{Method} & \textbf{Detector} & \textbf{Source-free} & \textbf{bus} & \textbf{bcycle} & \textbf{car} & \textbf{mcycle} & \textbf{person} & \textbf{rider} & \textbf{train} & \textbf{truck} & \textbf{mAP} \\
	\midrule 
	\textbf{Source only} & \multirow{8}{*}{Faster R-CNN} & & 26.0  & 29.7 & 35.8 & 22.4 & 29.3& 34.1 &9.10 &15.4 &25.2\\
  	SFOD-Mosaic (SED) \cite{li2021free} & & \checkmark & 39.0 & 34.1 & 44.5& 28.4& 33.2 &40.7 &22.2& 25.5 &33.5\\
   	LODS \cite{Li_2022_CVPR} & & \checkmark & \underline{39.7} & 37.8 & 48.8 & 33.2 & 34.0 & 45.7 & 19.6 & 27.3 &35.8\\
	IRG \cite{vibashan2023instance} & & \checkmark & 39.6 & \underline{41.6} & 51.9 & 31.5 & 37.4 & 45.2 & \underline{25.2} & 24.4 &\underline{37.1}\\
    A$^{2}$SFOD \cite{chu2023adversarial} & & \checkmark & 34.3 & 38.9 & 44.6 & 31.8 & 32.3 & 44.1 & \textbf{29.0} & \underline{28.1} & 35.4 \\
    RPL \cite{zhang2023refined} & & \checkmark & \textbf{46.7} & \textbf{46.5} & \underline{52.0} & \textbf{34.5} & \underline{37.9} & \textbf{52.7} & 21.7 & \textbf{29.9} & \textbf{40.2} \\
    PETS \cite{Liu2023PeriodicallyET} & & \checkmark & 39.3 & \underline{41.6} & \textbf{56.3} & \underline{34.2} & \textbf{42.0} & \underline{48.7} & 5.50 & 19.3 & 35.9 \\
 	\textbf{Oracle} & & & 49.4 & 38.8 & 56.7 & 35.5 & 38.7& 46.9 & 44.7 & 35.5 & 43.3\\ 
	\midrule
	\textbf{Source only} & \multirow{7}{*}{YOLOv5l}& & 35.7 & 44.8 & 64.1 & 32.7& 53.9 & 54.3& 22.8 & 26.0 &41.8\\
	SSDA-YOLO \cite{Zhou2022SSDAYOLOSD} & & \ding{53} & \textbf{63.0} & \textbf{53.6} & \textbf{74.3} & \textbf{47.4} & \textbf{60.6} & \textbf{62.1} & \underline{48.0} & \underline{37.8} & \textbf{55.9}\\
    YOLO-G \cite{wei2023yolo} & &  \ding{53} & 52.8 & 32.7 & 65.1 & \underline{44.8} & 46.0 & 47.5 & \textbf{55.1} & \textbf{38.2} & 47.8 \\
    PETS \cite{Liu2023PeriodicallyET} (our code)& &     \checkmark & 34.6 & 43.9 & 63.6 & 31.8 & 53.1 & 52.9 & 19.9 & 25.4 & 40.6 \\
    SF-YOLO (ours) & &              \checkmark & \underline{53.7} & 50.4 & 71.4 & 39.6 & 55.1 & \underline{58.2}  & 46.8 & 34.5  & 51.2 \\
 	SF-YOLO$^\dag$ (ours) & &              \checkmark & \underline{53.7} & \underline{50.5} & \underline{71.5} & 40.6 & \underline{55.5} & 58.0 & 46.1 & 36.6 & \underline{51.6} \\
	\textbf{Oracle} & & & 66.1 & 53.9 & 79.7 & 47.6 & 64.6 & 61.4 & 59.0 & 44.4 & 59.6\\
	\midrule
	\textbf{Source only} & \multirow{9}{*}{YOLOv5s}& & 31.9 & 31.5 & 51.3 & 13.3 & 39.7 & 38.8 & 2.20 & 12.5 &27.4\\
    S-DAYOLO \cite{Li2022CrossDomainOD} & & \ding{53} & 40.5 & 37.3 & 61.9 & 24.4 & 42.6 & 42.1 & \underline{39.5} & 23.5 & 39.0\\
 	SSDA-YOLO \cite{Zhou2022SSDAYOLOSD} & & \ding{53} & 45.6 & 38.8 & 53.8 & \textbf{34.3} & 43.8 & 44.9 & 34.7 & \textbf{27.3} & 40.4\\
    ConfMix \cite{Mattolin2022ConfMixUD} & & \ding{53} & 45.8 & 33.5 & 62.6 & \underline{28.6} & 45.0 & 43.4 & \textbf{40.0} & \textbf{27.3} & 40.8\\
   	CAST-YOLO \cite{liu2023cast} & & \ding{53} & 43.2 & \textbf{51.1} & \textbf{70.1} & \underline{28.6} & \textbf{54.0} & \textbf{58.9} & 19.2 & 21.4 & \textbf{43.3}\\
    PETS \cite{Liu2023PeriodicallyET} (our code)& &     \checkmark & 24.6 & 27.1 & 47.1 & 14.1 & 36.4 & 35.8 & 0.01 & 0.07 & 24.1 \\
   	SF-YOLO (ours) & &              \checkmark & \textbf{49.7} & 44.1 & 64.1 & 25.9 & 47.4 & 50.6  & 32.5 & \underline{26.0}  & 42.5 \\
    SF-YOLO$^\dag$ (ours) & &              \checkmark & \underline{48.1} & \underline{44.8}  & \underline{64.5}  & 23.1  & \underline{48.6}  & \underline{51.8}  & 37.6  & 24.7  & \underline{42.9} \\
	\textbf{Oracle} & & & 57.7 & 42.3 & 71.9 & 40.1 & 51.2 & 49.2 & 56.3 & 40.1 & 51.1\\
	\bottomrule
	\end{tabular}
	}
\end{table}

\begin{table}
	\centering
 \caption{AP accuracy of our SF-YOLO and state-of-the-art SFDA methods on the K2C and S2C scenarios. }
	\label{tab:combined}
	\resizebox{0.75\textwidth}{!}{%
	\begin{tabular}{
		p{0.35\textwidth}
		c
        @{\hspace{8pt}}
		c 
        @{\hspace{8pt}} | @{\hspace{10pt}}
		c 
        @{\hspace{18pt}}
		c
	}
	\toprule
	\multirow{2}{*}{\raisebox{-2ex}{\textbf{Method}}} & \multirow{2}{*}{\raisebox{-2ex}{\textbf{Detector}}} & \multirow{2}{*}{\raisebox{-2ex}{\textbf{Source-free}}} & \multicolumn{2}{c}{\textbf{Car AP}} \\[5pt]
	\cline{4-5}
	& & & \raisebox{-1ex}{\textbf{K2C}} & \raisebox{-1ex}{\textbf{S2C}} \\[5pt]
	\midrule 
	\textbf{Source Only} & \multirow{8}{*}{Faster R-CNN} & &36.4 & 33.7\\
   	SFOD-Mosaic (SED) \cite{li2021free} & & \checkmark & 44.6 & 42.9\\
 	LODS \cite{Li_2022_CVPR} & & \checkmark &43.9 & - \\
  	IRG \cite{vibashan2023instance} & & \checkmark & 45.7 & 43.2\\
    A$^{2}$SFOD \cite{chu2023adversarial} & & \checkmark & 44.9 & 44.0 \\
    RPL \cite{zhang2023refined} & & \checkmark & \textbf{47.8} & \underline{50.1} \\
    PETS \cite{Liu2023PeriodicallyET} & & \checkmark & \underline{47.0} & \textbf{57.8} \\
	\textbf{Oracle} & & & 58.5 & 58.5\\ 
	\midrule
	\textbf{Source only} & \multirow{6}{*}{YOLOv5l}& &59.8 & 63.9\\
	YOLO-G \cite{wei2023yolo} & & \ding{53} &62.8 & 64.2\\
    PETS \cite{Liu2023PeriodicallyET} (our code) & & \checkmark & 57.9 & 61.9 \\
	SF-YOLO (ours) & & \checkmark & \underline{62.7} & \underline{69.3}\\
    SF-YOLO$^\dag$ (ours) &  & \checkmark & \textbf{63.7}& \textbf{69.8}\\
	\textbf{Oracle} & & &84.6 & 84.6\\
	\midrule
	\textbf{Source only} & \multirow{8}{*}{YOLOv5s}& & 35.9& 48.8\\
    SimROD \cite{Ramamonjison2021SimRODAS} & & \ding{53} & 47.5 & 52.1\\
    S-DAYOLO \cite{Li2022CrossDomainOD} & & \ding{53} & 49.3 & - \\
    ConfMix \cite{Mattolin2022ConfMixUD} & & \ding{53} & \textbf{52.2} & \underline{56.3}\\
    PETS \cite{Liu2023PeriodicallyET} (our code) & & \checkmark & 33.0 & 47.6 \\
    SF-YOLO (ours) &  & \checkmark & 49.4 & \textbf{57.7}\\
    SF-YOLO$^\dag$ (ours) &  & \checkmark & \underline{50.6}& \textbf{57.7} \\
	\textbf{Oracle} & & & 70.3 & 70.3 \\
	\bottomrule
	\end{tabular}
	}
\end{table}

\begin{figure}[!htb]
  \centering
    \subfloat[Learning rate vs. SSM]{\includegraphics[width=0.3\linewidth]{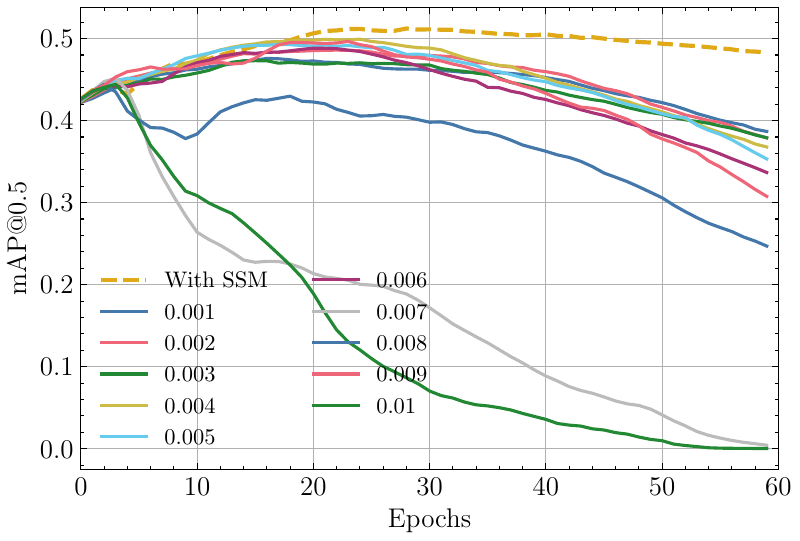} \label{fig:abl-lrVSssm}} \quad
    \subfloat[$L_2$ vs. SSM]
    {\includegraphics[width=0.3\linewidth]{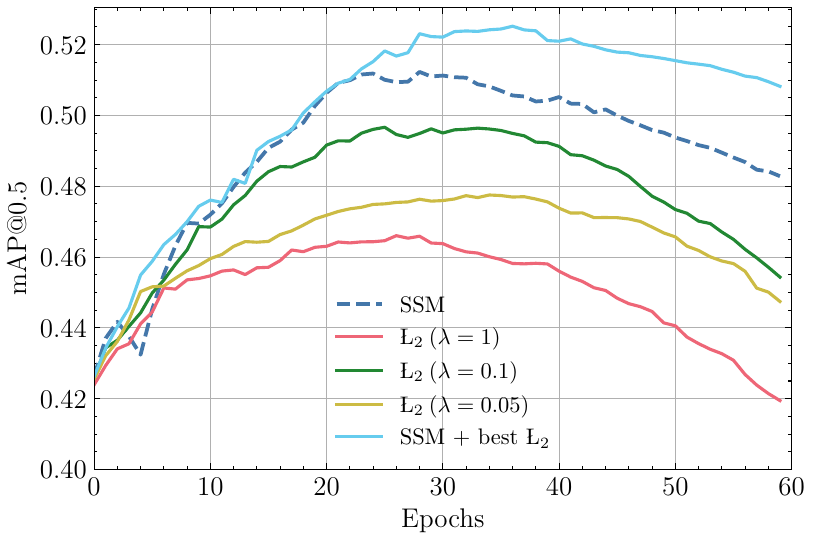} \label{fig:abl-l2}} \quad
    \subfloat[Delaying mean-teacher]{\includegraphics[width=0.3\linewidth]{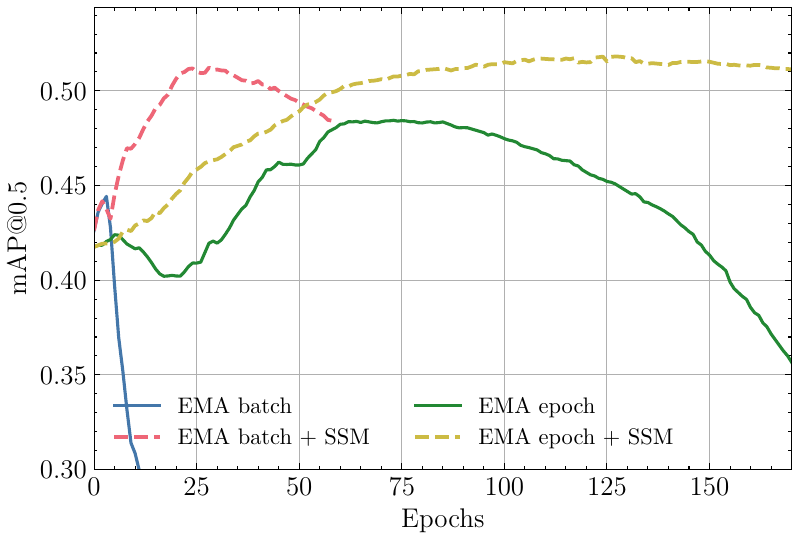} \label{fig:abl-ema}} \\

    \caption{The training curves of YOLOv5l on the C2F scenario for different mechanisms preventing drift between teacher and student outputs. All methods above can mitigate the drift, but SSM performs better in most cases without requiring extensive tuning.}
    \label{fig:ssm_yolov5l_mechanisms}
\end{figure}

\begin{figure}[!htb]
  \centering
  \subfloat[C2F]{\includegraphics[width=0.3\linewidth]{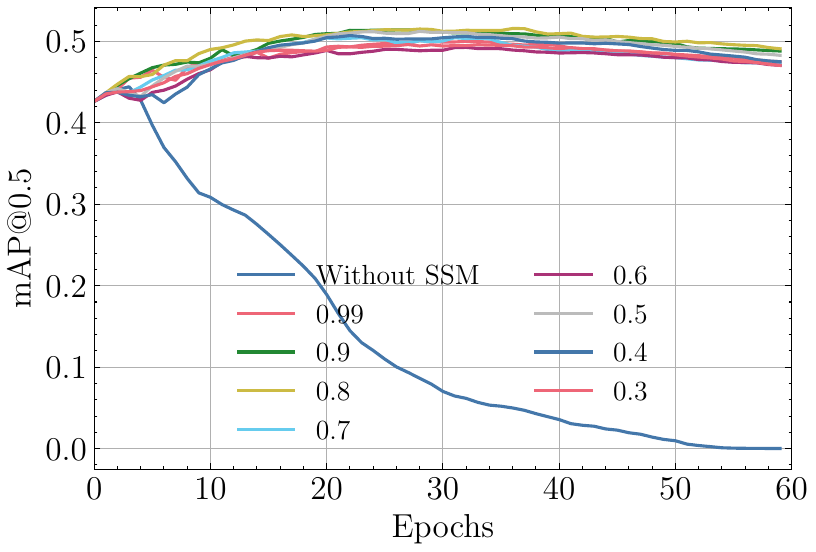} \label{fig:C2F_SSM}} \quad
  \subfloat[S2C]{\includegraphics[width=0.3\linewidth]{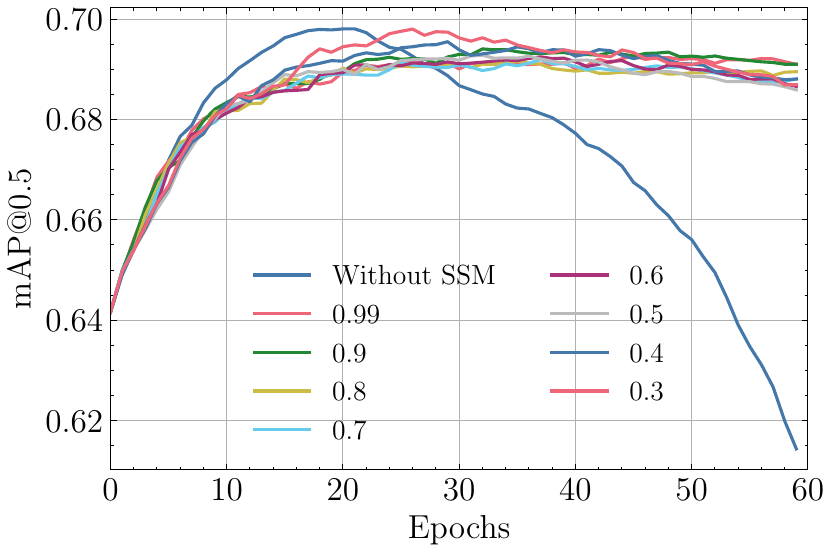} \label{fig:S2C_SSM}} \quad
  \subfloat[K2C]{\includegraphics[width=0.3\linewidth]{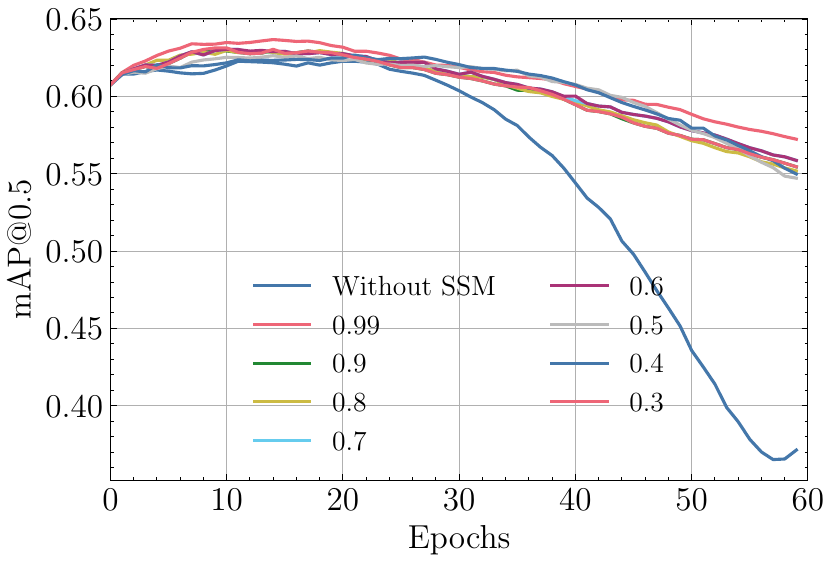} \label{fig:K2C_SSM}} \\
  \caption{The training curves of YOLOv5 on three scenarios using our method across a range of SSM momentum values. SSM produces relatively stable outcomes across different settings. For reference, we also show the results without SSM.}
  \label{fig:ssm_yolov5l}
\end{figure}

\subsection{Ablation Studies}

This section analyses the importance of TAM and SSM in the C2F scenario with YOLOv5l. It shows that learned augmentation offers greater benefits compared to random data augmentation. It also shows that SSM behaves differently from other regularization techniques used to control weight consistency between student and teacher. Moreover, alignment is found unnecessary in our setting, and SSM is compatible with other architectures, such as Faster R-CNN. For similar ablations with YOLOv5s, see the Suppl. Materials.

\noindent\textbf{(a) Learned Augmentation vs Random Augmentation.}
We compare the TAM with weak-strong random augmentations as described in \cite{lee2013pseudo}. Weak augmentations (image flipping, geometrical transformation) are applied to the images to be processed by the teacher model, while strong augmentations \cite{chen2020simple} (e.g., color jitter, randomize grayscale and blur) are applied to the images fed to the student model to improve the overall generalization capability. From \Cref{tab:augment} we see that although both techniques improve performance, learned augmentation provides higher gains since it is specific to the target dataset.

\begin{table}[htbp]
    \centering
    \caption{Ablation on TAM and Strong-Weak augmentation.}
    \resizebox{0.7\textwidth}{!}{
    \begin{tabular}{
    c
    | c
    | c
    }
    \toprule
    \textbf{Target augment.} & \textbf{Strong-Weak augment.} & \textbf{mAP (C2F)} \\
    \midrule
    \checkmark & \ding{53} & 51.2 \\
    \ding{53} & \checkmark & 41.4 \\
    \checkmark & \checkmark & 41.7 \\
    \bottomrule
    \end{tabular}}
    \label{tab:augment}
\end{table}

\noindent\textbf{(b) SSM vs Learning Rate.} One intuitive way to control the divergence between teacher and student is to restrict the learning speed of the student. In \Cref{fig:abl-lrVSssm}, we present the model performance over training iterations for our approach without SSM over a range of learning rates. As we can see, optimizing the learning rate indeed impacts the best possible outcome. Nonetheless, after the best performance, the accuracy degrades quickly, showing that the learning rate alone cannot control the drift between the student and the teacher model.

\noindent\textbf{(c) SSM vs $L_2$ between student and teacher.} Catastrophic forgetting is another perspective from which we can analyze the drift between the student and teacher model. Elastic weight consolidation (EWC) \cite{kirkpatrick2017overcoming} is a popular approach to overcome catastrophic forgetting in continual learning literature, where learning on certain weights is slower based on their importance on previously seen tasks. The original EWC requires the creation of a Fisher information matrix based on data from the source domain which cannot be satisfied in SFDA. Following Barone \etal\cite{barone2017regularization}, we use a simplified EWC version that approximates the Fisher information matrix as diagonal Gaussian prior with mean equal to the teacher parameters and $1/\lambda$ variance. The final penalty term is: 
\begin{equation}
	R_{L_2} = \lambda \cdot \lVert \phi - \theta \rVert_{2}^2
\end{equation}
and is simultaneously minimized with the detector loss. \Cref{fig:abl-l2} shows our experiments over different values of $\lambda$. First, we can see that the best performance is really sensitive to the choice of $\lambda$ and in some cases, a wrong value can reduce the performance by large a margin. Second, considering the best performing $\lambda$, SSM and $L_2$ have a complementary effect in which the overall best model is the combination of both. Nevertheless, the method used for the results in \Cref{tab:foggy} and \Cref{tab:combined} did not combine $L_2$ and SSM. Although this combination yielded the best performance, it required extensive hyperparameter tuning. In a real-world SFDA scenario where access to labels in target data is limited, conducting such tuning becomes impractical. Hence, we chose to exclude this combination in favor of a tuning-free SSM approach better suited to practical deployment.

\noindent\textbf{(d) Delaying mean-teacher.}
The graph depicted in \Cref{fig:abl-ema} displays the performance of our method without SSM under different $\text{EMA}_{\phi \rightarrow \theta}$ momentum values $\alpha$. Our intuition is that if the student updates the teacher too frequently, then the student does not have enough time to learn a robust representation of the target data, and this can introduce more noise to the teacher. Therefore, we have selected a small value for $\alpha$. By updating the teacher once every epoch instead of once every batch, we have observed that the training becomes more stable. However, similar to learning rate and $L_2$, the best performance is really sensitive to the choice of $\lambda$. Additionally, for the best performing $\lambda$, we can observe a complementary effect with SSM.

\noindent\textbf{(e) Hyper-parameters sensitivity.} Based on the results presented in \Cref{fig:ssm_yolov5l}, it can be observed that SSM produces relatively stable outcomes across a wide range of $\gamma$ values for different datasets. As anticipated, adjusting the gamma parameter leads to improved training stability and overall performance. However, regardless of the gamma value used, SSM provides training stabilization and generally outperforms the default MT setting. This highlights the significance of tuning-free SSM in achieving near-optimal performance in SFDA.

\noindent\textbf{(f) Feature alignment.} Feature alignment of the source and target domains is typically used in UDA methods to improve generalization by extracting domain-invariant features \cite{chen2018domain,Saito2018StrongWeakDA,Hsu2020EveryPM,Wang2021AFANAF,Zhuang2020iFANIF}. Although SFDA doesn't allow direct feature alignment due to the absence of source domain data, our SF-YOLO framework uses a TAM. This module enables the creation of a pseudo-domain $D_{aug} = (\mathcal{X}_{aug}) = \{\text{TAM}(x_t^i) \}_{i=1}^{N_t}$ representing an augmented version of the target domain $D_t$. Our findings indicate that aligning target domain features with their augmented counterparts does not enhance performance in our SFDA setting. As a result, we opted against using feature alignment to keep our method simpler. For more details and the complete results, please refer to the Suppl. Material.

\noindent\textbf{(g) Faster R-CNN.} To show the compatibility of SSM with other architectures than YOLO, we applied SSM to IRG \cite{vibashan2023instance}, one of the SOTA methods using Faster R-CNN architecture. Table \ref{tab:faster-rcnn-irg} shows the compatibility of SSM with Faster R-CNN. Utilizing SSM provides a significant gain in performance in all scenarios. It allows to achieve SOTA accuracy for SFDA methods based on Faster R-CNN on K2C, and second and third best on C2F and S2C scenarios, respectively.  
While the aforementioned methods can reduce drifting, SSM outperforms in most cases and requires minimal fine-tuning, which is crucial for SFDA. Furthermore, we find feature alignment to be unnecessary in our setting, and our proposed SSM is compatible with other architectures, like Faster R-CNN.

\begin{table}[!t]
    \centering
    \caption{mAP accuracy of SFDA Faster R-CNN based methods and IRG with and w/wo SSM on the C2F, K2C and S2C scenarios.}
    \begingroup
    \setlength{\tabcolsep}{10pt}
    \resizebox{1\textwidth}{!}{
    \begin{tabular}{
    l
    | c
    | c
    | c
    | c
    | c
    }
    \toprule
    \textbf{Method} & \textbf{Venue} & \textbf{Detector} & \textbf{C2F} & \textbf{K2C} & \textbf{S2C} \\
    \midrule
    \textbf{Source only} & NIPS 2015 & Faster R-CNN & 25.2 & 36.4 & 33.7 \\
    SFOD-Mosaic (SED) \cite{li2021free} & AAAI 2021 & Faster R-CNN & 33.5 & 44.6 & 42.9 \\
    LODS \cite{Li_2022_CVPR} & CVPR 2022 & Faster R-CNN & 35.8 & 43.9 & - \\
    A$^2$SFOD \cite{chu2023adversarial} & AAAI 2023 & Faster R-CNN & 35.4 & 44.9 & 44.0 \\
    RPL \cite{zhang2023refined} & ICASSP 2023 & Faster R-CNN & \textbf{40.2} & \underline{47.8} & \underline{50.1} \\
    PETS \cite{Liu2023PeriodicallyET} & ICCV 2023 & Faster R-CNN & 35.9 & 47.0 & \textbf{57.8} \\
    \midrule
    IRG \cite{vibashan2023instance} & CVPR 2023 & Faster R-CNN & 37.1 & 45.7 & 43.2 \\
    IRG + SSM (ours) & - & Faster R-CNN & \underline{38.6} & \textbf{50.5} & 48.9 \\
    \midrule
    \textbf{Oracle} & NIPS 2015 & Faster R-CNN & 43.3 & 58.5 & 58.5 \\
    \bottomrule
    \end{tabular}}
    \endgroup
    \label{tab:faster-rcnn-irg}
\end{table}

\section{Conclusion}

We proposed the first approach for SFDA using the YOLO family of single-shot detectors. Our method employs a teacher-student framework with a learned, target domain-specific augmentation and a novel communication mechanism to stabilize training, reducing reliance on annotated target data for model selection, which is crucial for real-world applications. SF-YOLO outperforms all SFDA methods based on Faster R-CNN and even some UDA YOLO-based methods using source data. We present SSM, a simple addition to the MT framework, which improves performance and stability. Our extensive experiments show SSM's compatibility with existing knowledge preservation techniques. Our primary hypothesis behind the success of SSM lies in the online distillation of knowledge between student and teacher, which prevents the drift of models caused by noisy pseudo-labels. SSM is particularly well-suited for source-free (unsupervised) learning scenarios. Unlike conventional semi-/weakly-/unsupervised learning, these scenarios have no access to source data nor the labeled data from the target domain to ensure the stability of the model and prevent significant drift during training.

\section*{Acknowledgements}
This work was supported by Distech Controls Inc., the Natural Sciences and Engineering Research Council of Canada, and the Digital Research Alliance of Canada. 

\clearpage

\title{Supplementary Material: Source-Free Domain Adaptation for YOLO Object Detection} 

\titlerunning{Suppl. Material: SFDA for YOLO Object Detection }

\author{Simon Varailhon \and
Masih Aminbeidokhti\orcidlink{0000-0003-2289-3690} \and
Marco Pedersoli\orcidlink{0000-0002-7601-8640} \and
Eric Granger\orcidlink{0000-0001-6116-7945}} 

\authorrunning{S.~Varailhon et al.}

\institute{LIVIA, ILLS, Dept. of Systems Engineering, ETS Montreal, Canada \\
\email{\{simon.varailhon.1, masih.aminbeidokhti.1\}@ens.etsmtl.ca\\ \{marco.pedersoli, eric.granger\}@etsmtl.ca}}

\maketitle

In the supplementary material, \Cref{sec:datasets} provides more details on the adaptation scenarios and related datasets considered in our experiments. Furthermore, to show that our proposed method works across different model sizes, \Cref{sec:yolov5s} provides similar experiments from the main paper but with YOLOv5s -- a smaller version of the YOLOv5 model. Finally, \Cref{sec:align} provides the detailed results of our feature alignment experiments. Our code is available at \url{https://github.com/vs-cv/sf-yolo}

\section{Adaptation Scenarios}
\label{sec:datasets}
\subsection{Datasets:}

The following four datasets were used for our experiments:\\
\noindent  \textbf{- Cityscapes} \cite{Cordts2016TheCD} is a collection of urban street scenes gathered across 50 cities over several months. The dataset is composed of images captured during daylight and favorable weather conditions. It provides annotations for eight categories: bus, bicycle, car, motorcycle, person, rider, train, and truck. With 2975 images in the training set and 500 in the validation set, it represents a comprehensive benchmark for evaluating OD algorithms in urban contexts, particularly useful for applications in autonomous vehicles.\\
\noindent  \textbf{- Foggy Cityscapes} \cite{Sakaridis2017SemanticFS} addresses the challenge of OD in adverse weather scenarios. It is created by simulating fog on top of Cityscapes images. As in previous works \cite{Zhou2022SSDAYOLOSD,Li_2022_CVPR}, for our experiments, we use the foggy level of 0.02 which corresponds to a visibility of 150 meters.\\
\noindent  \textbf{- KITTI} \cite{geiger2012we} is a popular dataset in autonomous driving scenarios and mobile robotics. The training set for 2D OD includes 7481 real labeled images. It is similar to the Cityscapes dataset but with different camera modalities. \\
\noindent \textbf{- Sim10k} \cite{johnson2016driving} is a synthetic dataset created using the game engine of Grand Theft Auto V. It is made of 10,000 images, all featuring cars in a variety of different environments and situations. It is used for domain adaptation in the context of methods developed that leverage this inexpensive synthetic data to (pre)train models for real-world applications.

\begin{figure}
  \centering
  \subfloat[Cityscapes \cite{Cordts2016TheCD}]{\includegraphics[height=2.2cm]{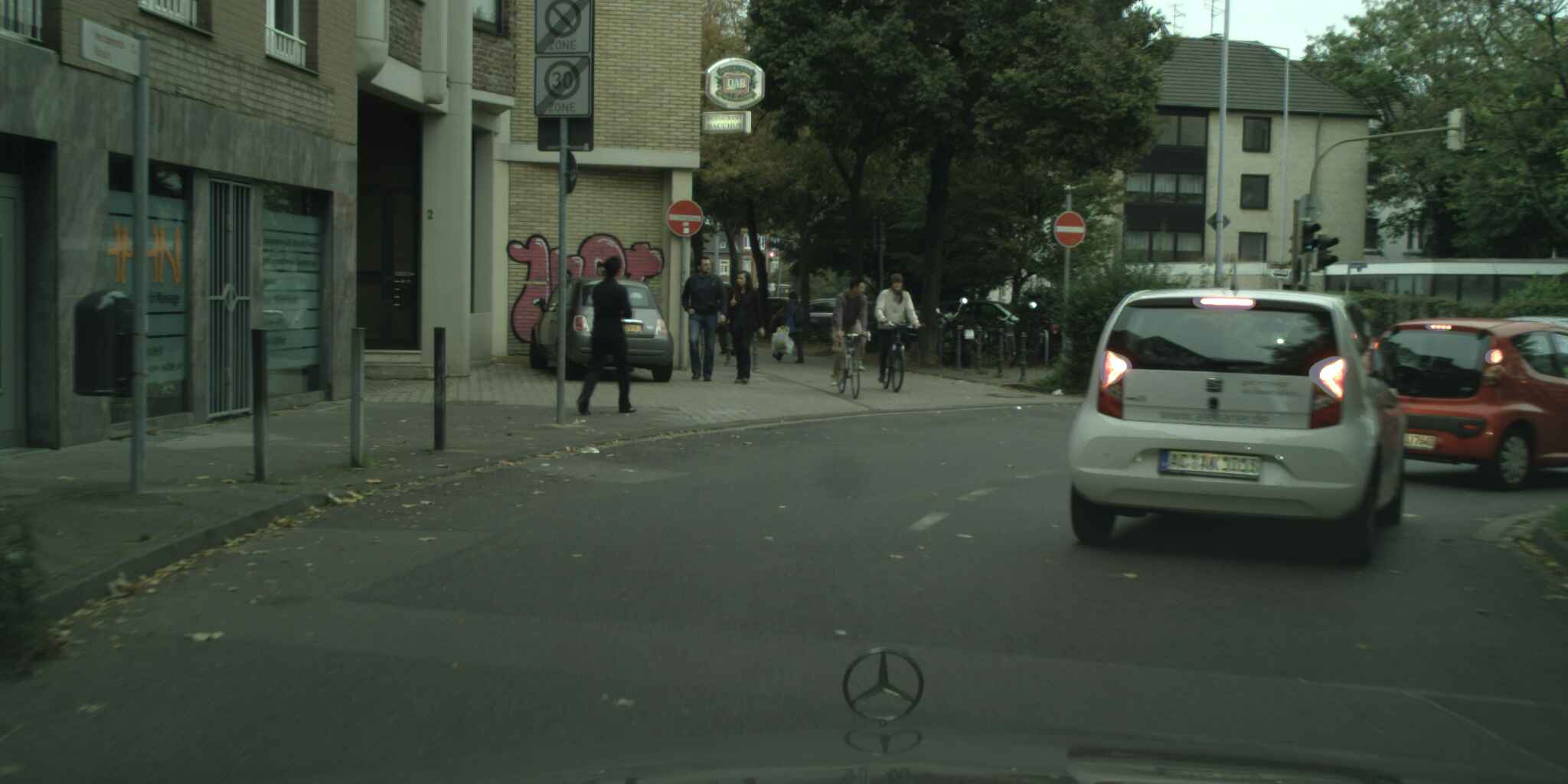} \label{fig:cityscapes}} \quad
  \subfloat[Foggy Cityscapes \cite{Sakaridis2017SemanticFS}]{\includegraphics[height=2.2cm]{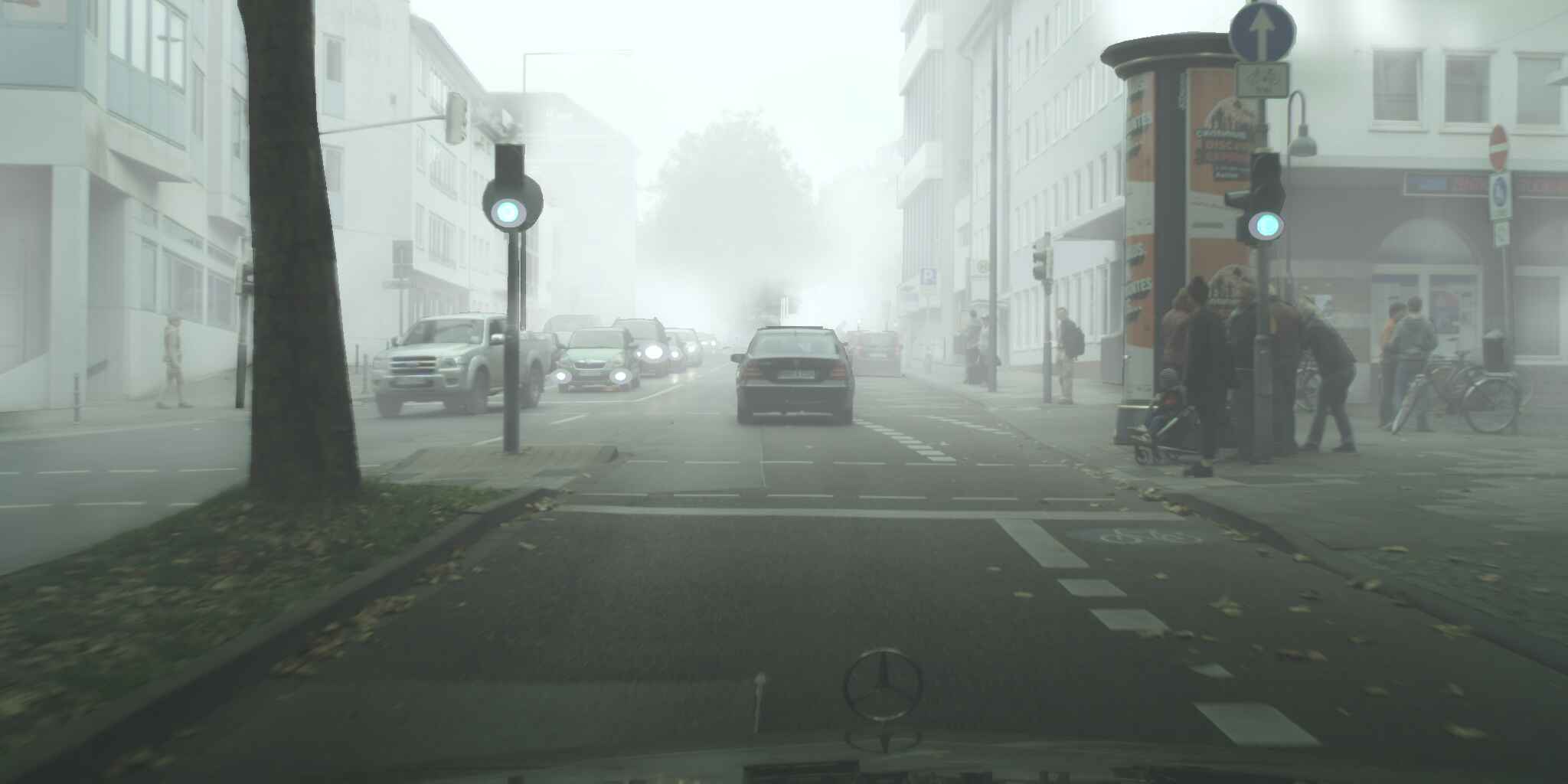} \label{fig:foggy}} \quad
  \subfloat[Sim10k \cite{johnson2016driving}]{\includegraphics[height=2.2cm]{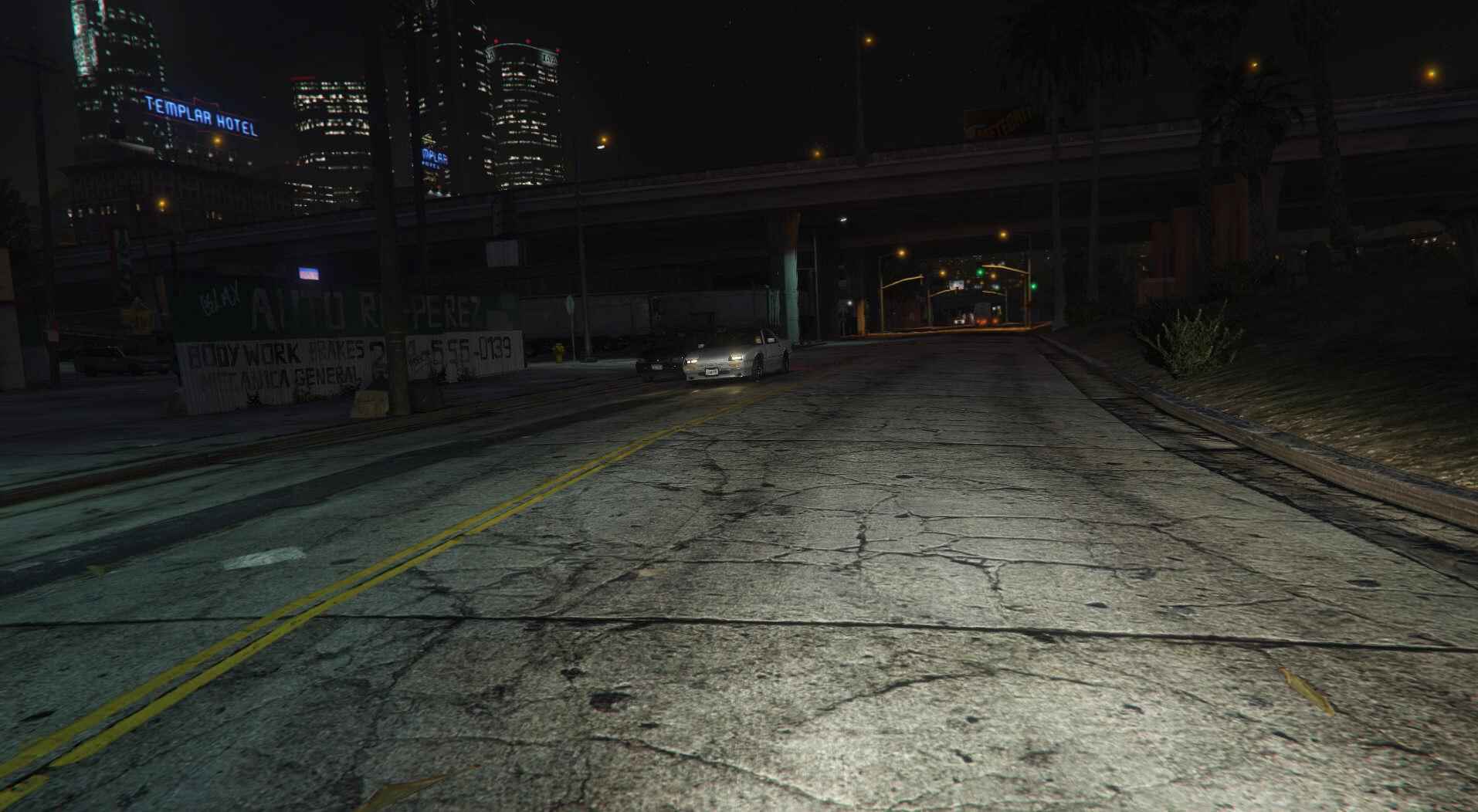} \label{fig:sim10k}} \\
  \subfloat[KITTI \cite{geiger2012we}]{\includegraphics[height=2.2cm]{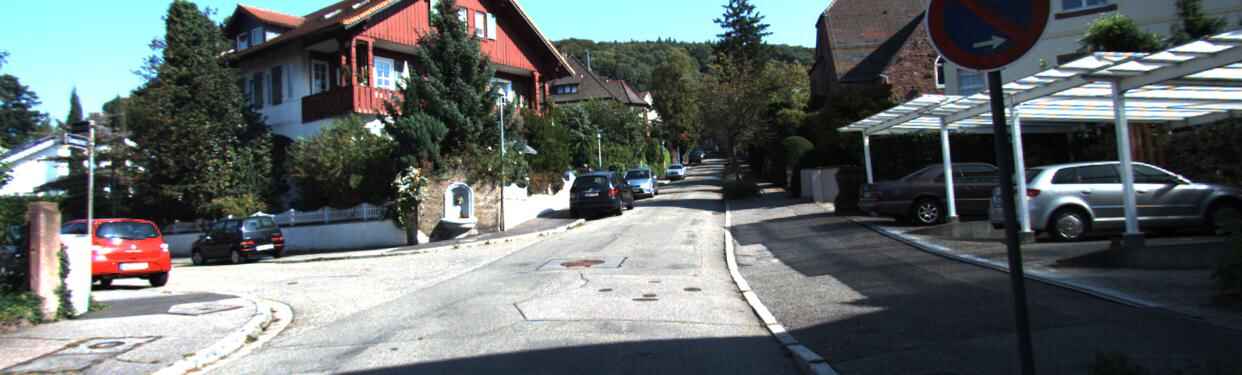} \label{fig:kitti}} \\
  \caption{Selected sample from the four datasets used in our experiments.}
  \label{fig:ssm_yolov5s}
\end{figure}

\subsection{Domain shift:}
The domain shift was analyzed across the domain adaptation scenarios used in our experiments. To do this, we utilized a pretrained model on the source domain of each scenario. Then, we extracted features from both source and target domain datasets and calculated the domain shift using Maximum Mean Discrepancy (MMD) at the feature level. In the scenario of \textit{Cityscapes $\rightarrow$ Foggy Cityscapes} (C2F), we observed the smallest MMD value at 3.80, which is reasonable given the similarity between the domains; the images are identical except for synthetic fog added to the target images. In contrast, the domain shift is larger for \textit{KITTI $\rightarrow$ Cityscapes} (K2C) with an MMD value of 24.54 due to differences in camera settings and environments between the two domains. However, these are still real images, resulting in a smaller domain shift compared to the adaptation from synthetic images to a real-world dataset in the case of \textit{Sim10K $\rightarrow$ Cityscapes} (S2C), where the highest MMD value of 49.79 is recorded.

\section{Ablation Studies with YOLOv5s}
\label{sec:yolov5s}

In this section, we conduct similar experiments as within our paper but with YOLOv5s, a smaller version of the YOLOv5l model. Our findings reveal that the SSM enhances the stability and generally outperforms models trained without it, as shown in \Cref{fig:ssm_yolov5s}. The learning rate impacts the performance, but the same learning rate with and without SSM exhibits distinct trends, indicating that the SSM impact on performance cannot be replicated by adjusting the learning rate alone (\Cref{fig:abl-lrVSssm,fig:abl-samelrssm}). To stabilize the training, we found that updating the teacher once every epoch instead of once every batch improves the performance, although it is slower. However, coupling SSM with this method further enhances the performance (\Cref{fig:yolov5s_EMA_epochVSbatch}). Lastly, \Cref{fig:ssm_yolov5l} illustrates that SSM produces relatively more stable outcomes across a wide range of $\gamma$ values for different datasets. Overall, our results for YOLOv5s are similar to those obtained for YOLOv5l, confirming the effectiveness of the SSM in improving the model's performance and stability across different model sizes.

\begin{figure}
  \centering
  \subfloat[$\eta$ = 0.01]{\includegraphics[width=0.3\linewidth]{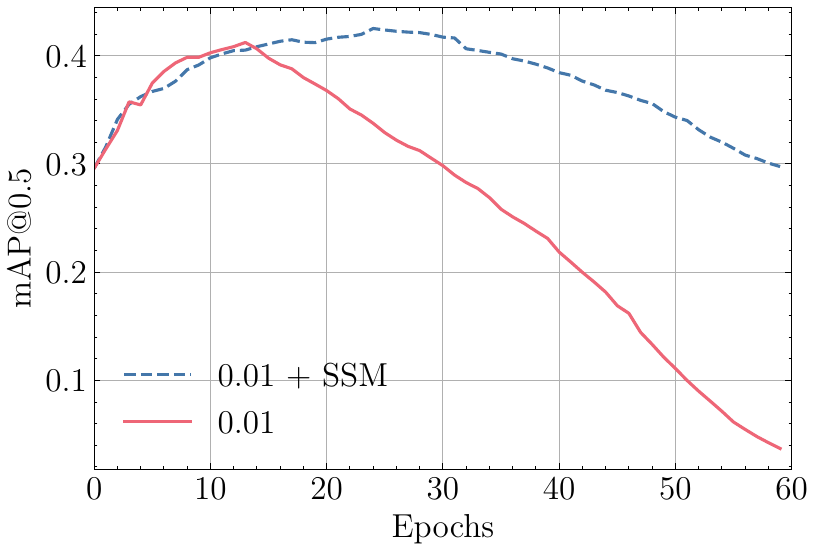} \label{fig:lr0}} \quad
  \subfloat[$\eta$ = 0.008]{\includegraphics[width=0.3\linewidth]{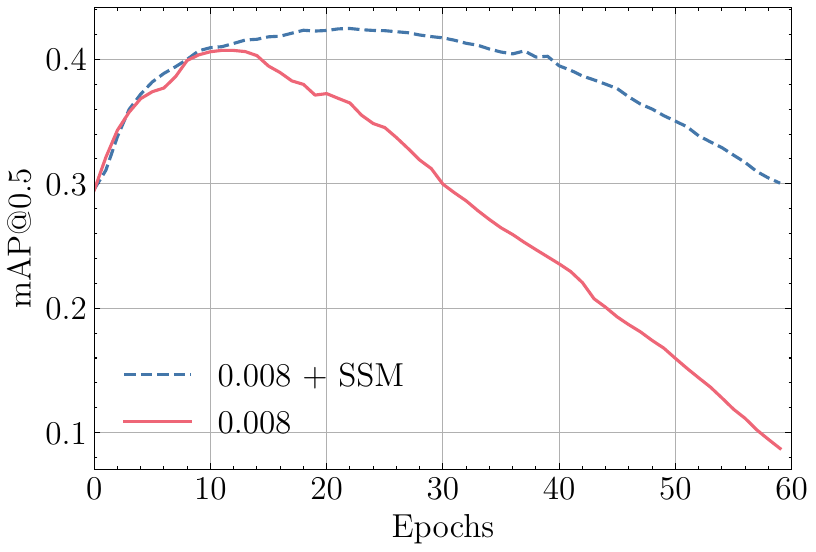} \label{fig:lr1}} \quad
  \subfloat[$\eta$ = 0.006]{\includegraphics[width=0.3\linewidth]{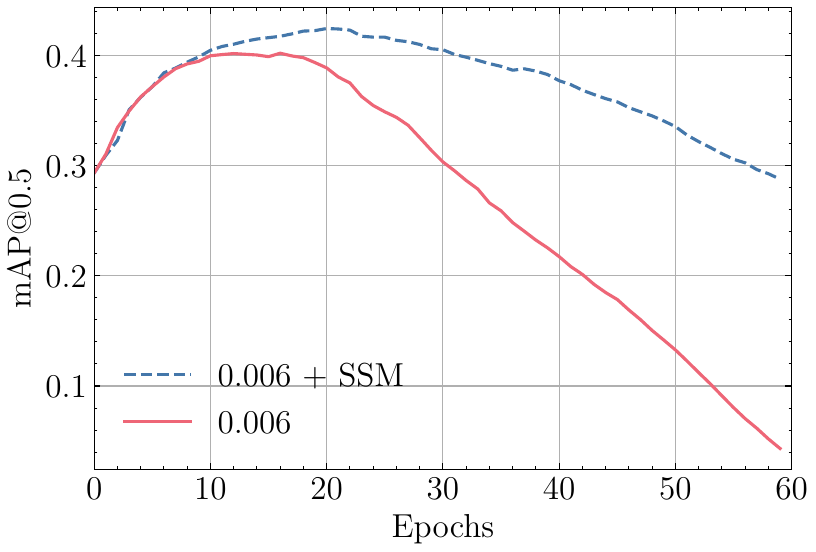} \label{fig:lr2}} \\
  \caption{The training curves of the target augmented mean-teacher with and without SSM using YOLOv5s on the C2F scenario across different learning rates. SSM prevents MT from quick deterioration and reaches better final performance.}
  \label{fig:ssm_yolov5s}
\end{figure}

\begin{figure}
  \centering
  \subfloat[Learning rate vs. SSM]{\includegraphics[width=0.3\linewidth]{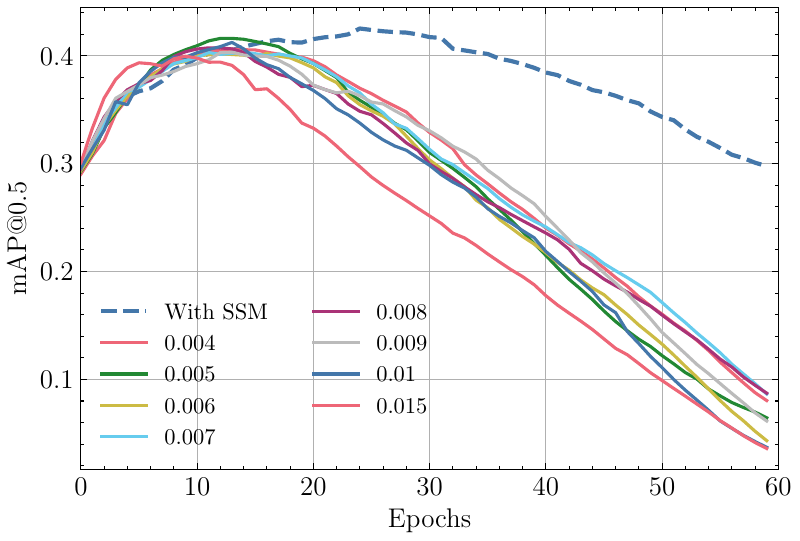} \label{fig:abl-lrVSssm}} \quad
  \subfloat[Same learning rate (within the 0.004 to 0.015 range) with and without SSM. Individual legends are omitted for clarity to highlight the overall trend but are detailed in Figure \ref{fig:abl-lrVSssm}]{\includegraphics[width=0.3\linewidth]{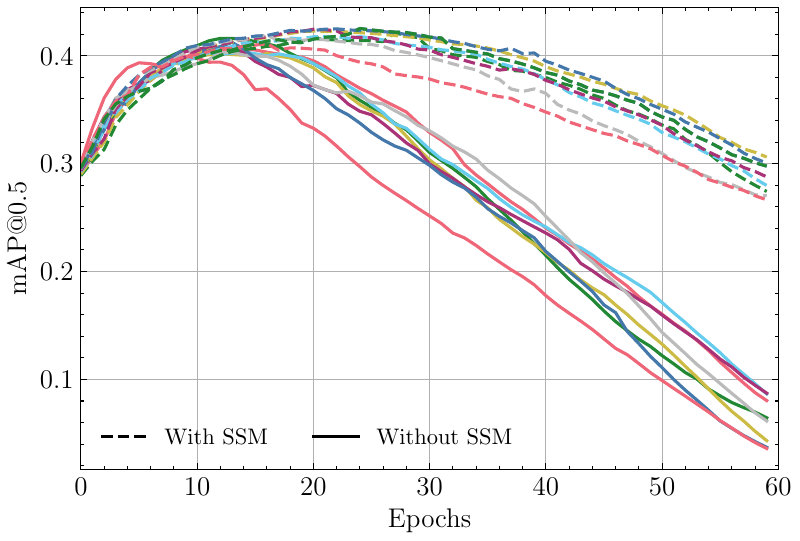} \label{fig:abl-samelrssm}} \quad
  \subfloat[Delaying mean-teacher]{\includegraphics[width=0.3\linewidth]{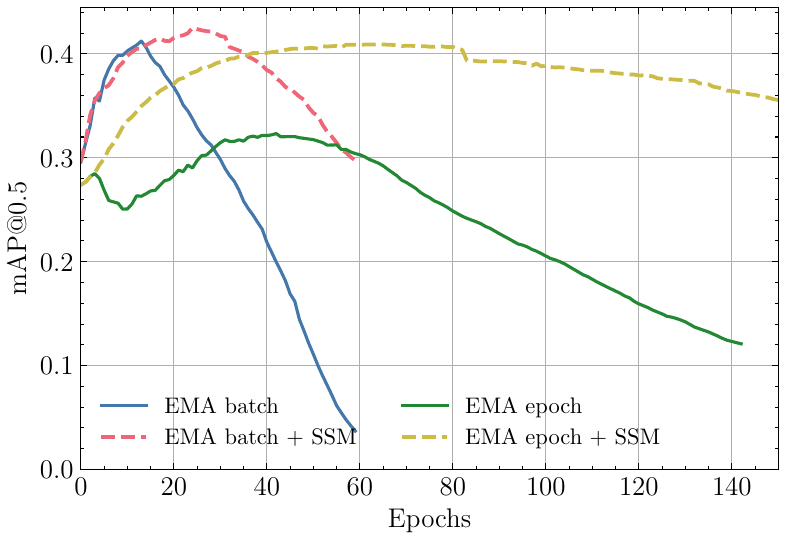} \label{fig:yolov5s_EMA_epochVSbatch}} \\
  \caption{The training curves of YOLOv5s on the C2F scenario for different mechanisms preventing drift between teacher and student outputs. All of the methods can mitigate drift but SSM still performs better in most cases without requiring extensive tuning.}
  \label{fig:ssm_yolov5l}
\end{figure}

\begin{figure}
  \centering
  \subfloat[C2F]{\includegraphics[width=0.3\linewidth]{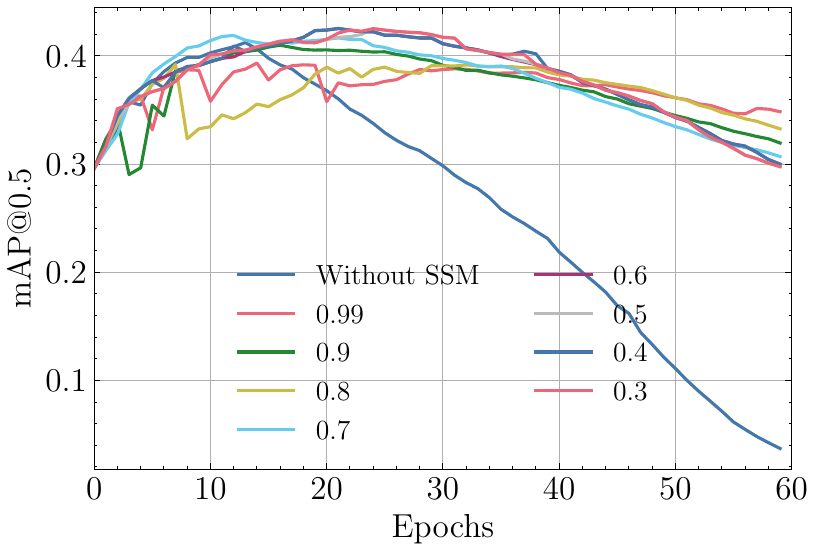} \label{fig:C2F}} \quad
  \subfloat[S2C]{\includegraphics[width=0.3\linewidth]{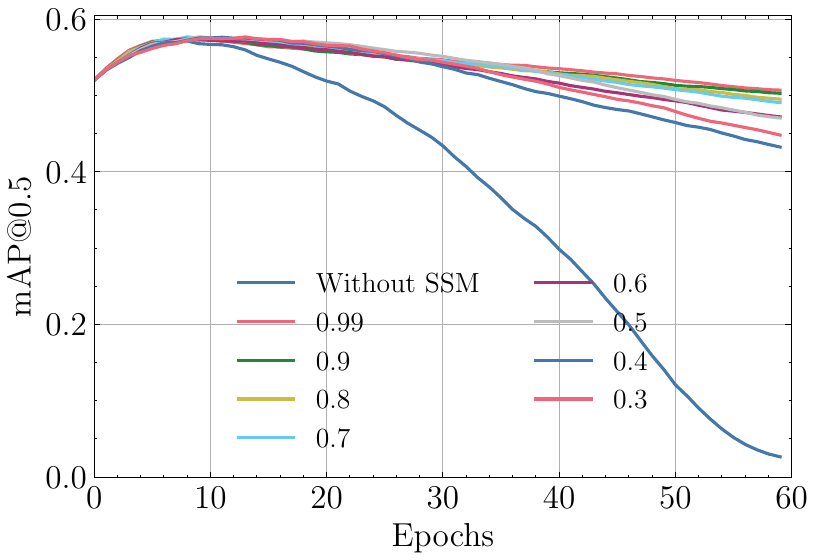} \label{fig:S2C}} \quad
  \subfloat[K2C]{\includegraphics[width=0.3\linewidth]{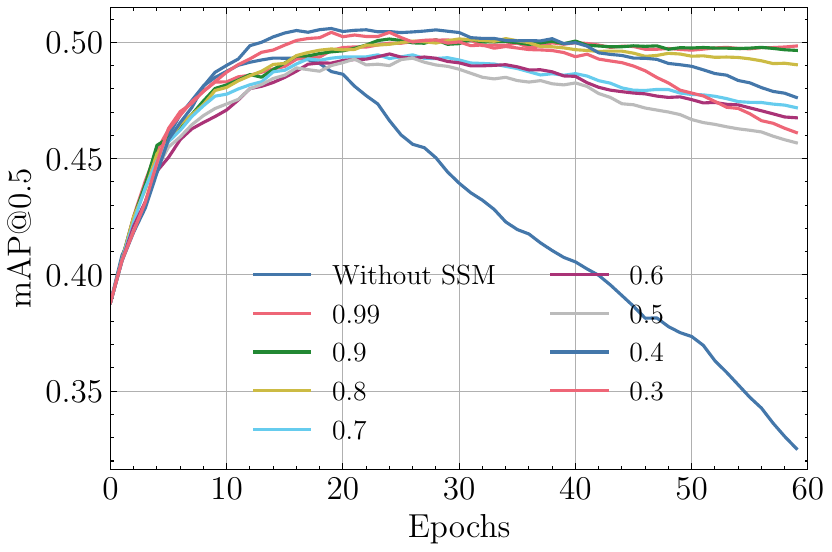} \label{fig:K2C}} \\
  \caption{The training curves of YOLOv5s on three scenarios using our method across a range of SSM momentums ($\gamma \in [0.3,0.99]$). SSM produces relatively stable outcomes across different settings. For reference, we also show the results without SSM.}
  \label{fig:ssm_yolov5l}
\end{figure}

\section{Feature Alignment}
\label{sec:align}

In this section, the feature alignment mechanisms are analyzed to assess their impact on the SF-YOLO method presented in the main manuscript. 

In the SFDA paradigm, direct feature alignment between domains is challenging due to the unavailability of source domain data. However, our SF-YOLO framework addresses this limitation by incorporating a Target Augmentation Module (TAM), as detailed in Section 3 of the main manuscript. This module enables the creation of a pseudo-domain $D_{aug} = (\mathcal{X}_{aug}) = \{\text{TAM}(x_t^i) \}_{i=1}^{N_t}$ representing an augmented version of the target domain $D_t = (\mathcal{X}_{t})$ . Our objective is to explore whether aligning features between these two domains enhances domain adaptation performance on target domain images.

Using a mean-teacher architecture, there are various ways to perform alignment. For instance, features extracted by the teacher and the student can be aligned such as in LODS \cite{Li_2022_CVPR}, or the features extracted by the student can be directly aligned as done by A$^2$SFOD authors \cite{chu2023adversarial}. For the latter, a slight modification to the SF-YOLO architecture is required, as in the original one, the student only sees augmented target images $\mathcal{X}_{aug}$. To align features, target images $\mathcal{X}_{t}$ are also provided to the student in this case. 

\begin{figure}
	\centering
	\includegraphics[width=0.97\textwidth]{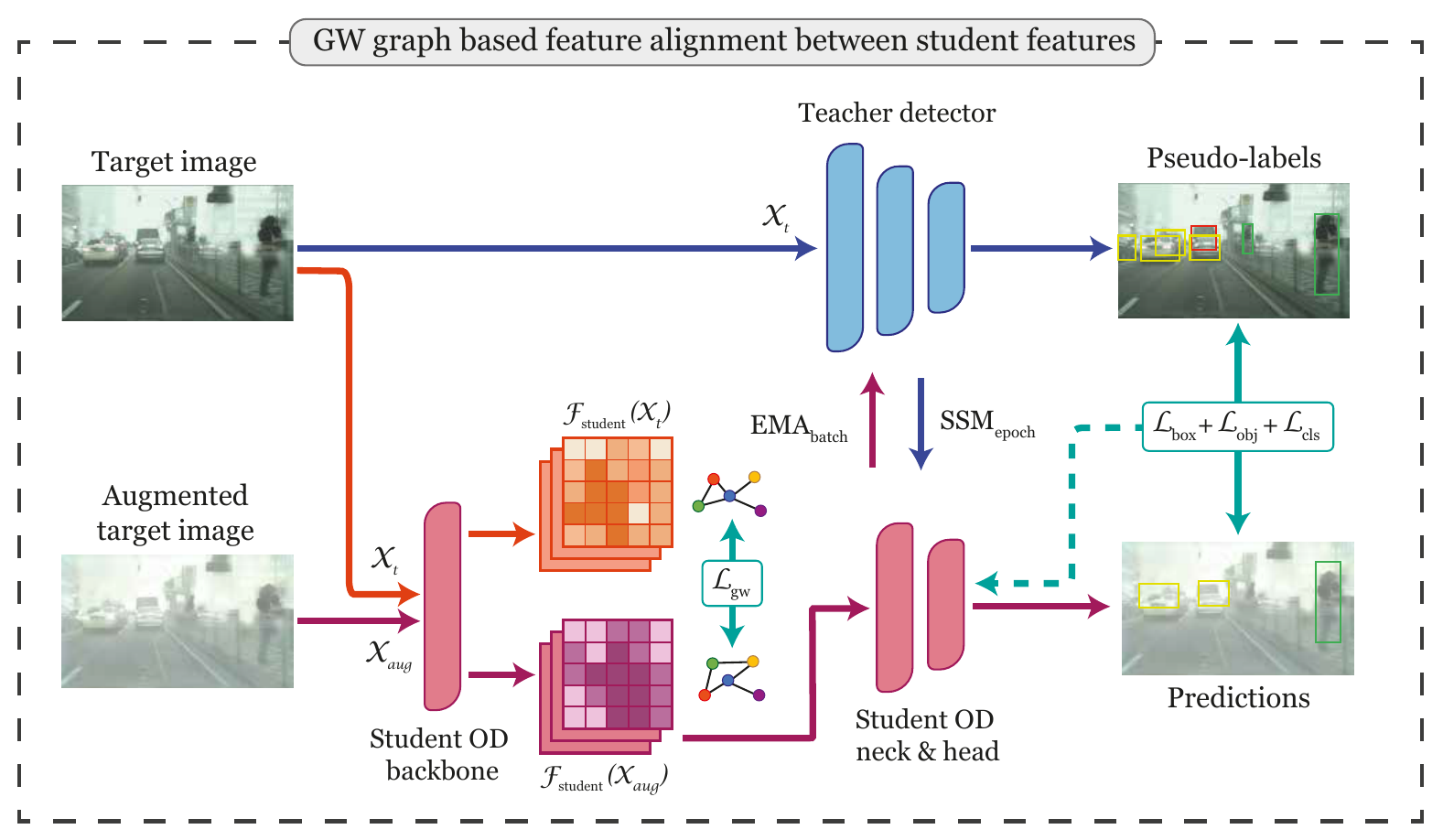}
	 \\ \parbox{1\textwidth}{\caption{GW graph-based feature alignment the between student features. The first data augmentation part of SF-YOLO (Target Augmentation Module) remains the same and is not represented in this figure. However, for this setting, the original SF-YOLO framework is modified so the student OD model receives both the target images and the augmented ones to allow the alignment process at the student model's level. 
  }\label{fig:gw_align_student}}
\end{figure}

\begin{figure}
	\centering
	\includegraphics[width=0.97\textwidth]{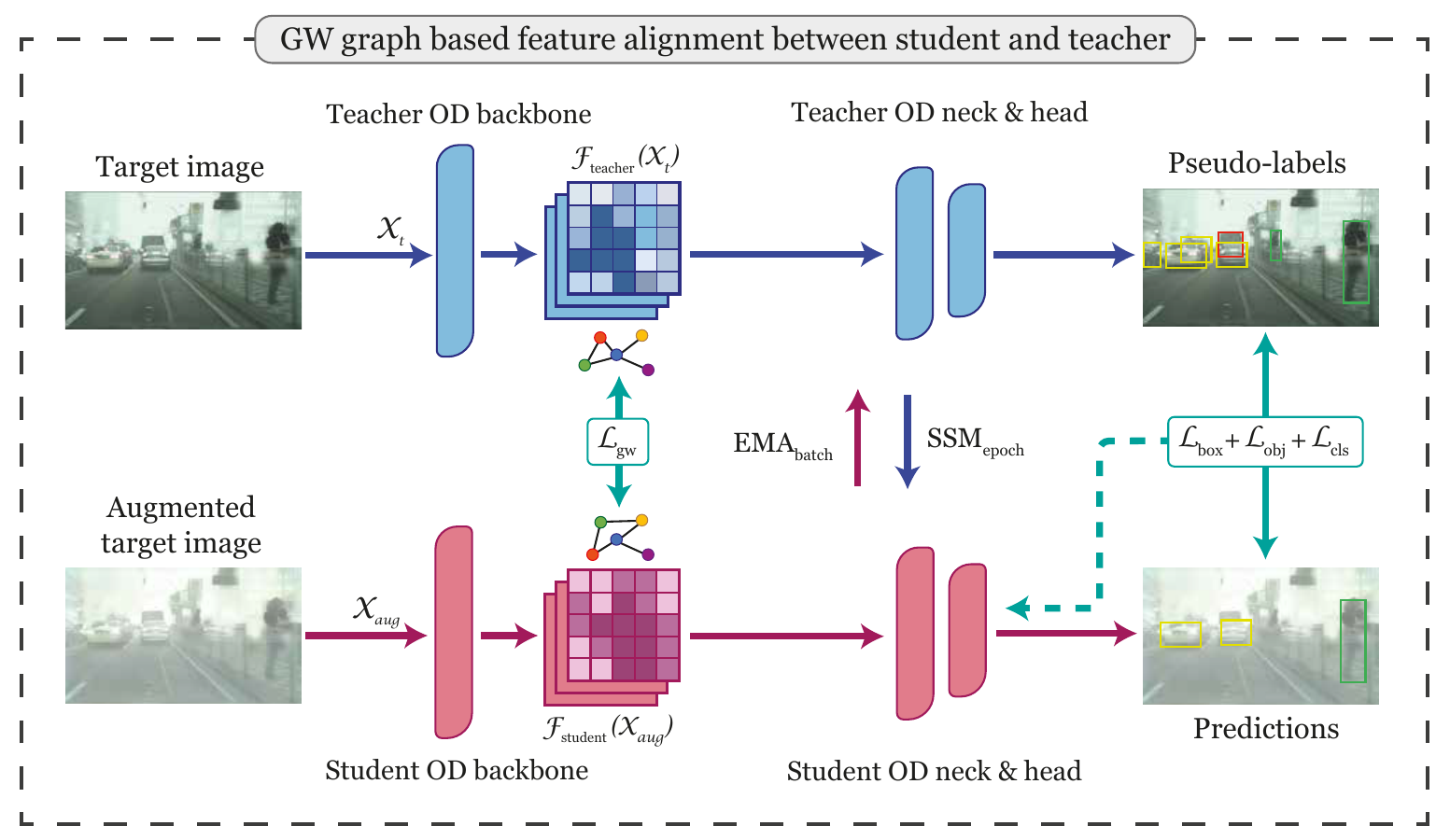}
	 \\ \parbox{1\textwidth}{\caption{GW graph-based feature alignment between the student and teacher features. The Target Augmentation Module is not represented in this figure but the overall architecture of SF-YOLO remains the same in this setting, the only difference being the GW alignment between the student and teacher backbones features.
  }\label{fig:gw_align_teacher_student}}
\end{figure}

The study is conducted in two parts :
\begin{itemize}
    \item \textbf{Alignment of student features}: The first part of the study focuses on aligning the features extracted by the student detector backbone from both the original and augmented target inputs. These features are denoted as \(\mathcal{F}_{\text{student}}(\mathcal{X}_{t})\) and \(\mathcal{F}_{\text{student}}(\mathcal{X}_{aug})\) respectively. The alignment is done using two different methods: Gromov-Wasserstein (GW) graph-based feature alignment as shown in \Cref{fig:gw_align_student} and adversarial feature alignment as depicted in \Cref{fig:ADV_align_student}.
    \item \textbf{Teacher-student feature alignment}: The study then investigates the impact of aligning the features extracted by the teacher \(\mathcal{F}_{\text{teacher}}(\mathcal{X}_{t})\) and the student \(\mathcal{F}_{\text{student}}(\mathcal{X}_{aug})\), this time using the unmodified SF-YOLO architecture as described in Figure 1 of the main manuscript. The alignment between features is done by the GW graph-based feature alignment method and is illustrated in \Cref{fig:gw_align_teacher_student}. The aim is to understand to what extent the alignment between the teacher and student features affects the domain adaptation learning process. 
\end{itemize}

\subsection{Gromov-Wasserstein graph-based feature alignment}

Similar to the work of \cite{Li_2022_CVPR}, we implement a GW  \cite{peyre2016gromov} discrepancy graph-based feature alignment strategy. Indeed, GW optimal transport has recently gained attention as an effective method for comparing and aligning similarity structures in an unsupervised manner \cite{kawakita2024gromov}. It has shown success in various applications, including the alignment of word embedding spaces across different languages \cite{alvarez2018gromov} and the matching of image features in domain adaptation \cite{Li_2022_CVPR}. This method is particularly good at revealing the underlying structural relationships in data without requiring supervision, which is especially relevant for the graph structures used in this section. The objective is to align two images features \(\mathcal{F}(\mathcal{X}_{t})\in \mathbb{R}^{H\times W\times C}\) and \(\mathcal{F}({X}_{aug})\in \mathbb{R}^{H\times W\times C}\), where $H$ and $W$ correspond to the dimensions of the feature maps and $C$ their number of channels. The features are either extracted from the teacher or the student models backbones $\mathcal{F}$ depending on the experiment. For this, \(\mathcal{F}({X}_{t})\) and \(\mathcal{F}(\mathcal{X}_{aug})\) are divided into $H\times W$ feature patches nodes $P_t$ and $P_{aug}$. By connecting all nodes with a weighting based on cosine similarity, two graphs $G_t(P_t,E_t)$ and $G_{aug}(P_{aug},E_{aug})$ are constructed, where $E_t$ and $E_{aug}$ are the edge-adjacency matrices of those features patches, which corresponds here to their cosine similarity matrices as follows:
\begin{equation}
 \label{eq:cosine}
    E_t = \frac{P_t \cdot P_t^\top }{\lVert P_t\lVert \cdot \lVert  P_t^\top \lVert}.
\end{equation}

The GW alignment loss is obtained by computing the GW distance as per Eq. \ref{eq:gw}:
\begin{equation}
    \label{eq:gw}
    \mathcal{L}_{gw} = \sum_{i ,i',j ,j'}KL(E_t^{i,i'}\lVert E_{aug}^{j,j'})T_{i,j}T_{j,j'},
\end{equation}
where $KL(\cdot \lVert \cdot)$ is the Kullback-Leibler divergence, measuring the distance of the edges across the graph and $T$ the graph matching matrix, representing the coupling between each feature of each node
of each graph, here equals to $I$ as the alignment objective is to ensure the correspondence of the features from the same patch.

Finally, the total loss of the SF-YOLO framework with GW feature alignment is given by:

\begin{equation}
    \mathcal{L}_{\text{det}} = \lambda_{\text{b}} \mathcal{L}_{\text{box}} + \lambda_{\text{c}} \mathcal{L}_{\text{cls}} + \lambda_{\text{d}} \mathcal{L}_{\text{obj}} + \lambda_{\text{g}} \mathcal{L}_{\text{gw}} \ , 
    \label{eq:yolo_loss_gw}
\end{equation}
\noindent where $\mathcal{L}_{\text{box}}$ and $\mathcal{L}_{\text{cls}}$ respectively represent the classification and bounding box regression losses, $\mathcal{L}_{\text{obj}}$ corresponds to the objectiveness loss related to the confidence of object presence, and $\mathcal{L}_{\text{gw}}$ denotes the GW loss. The $\lambda$ terms are weighting hyperparameters that control the relative importance of each loss component in the overall detection loss.

\subsection{Adversarial-based feature alignment}

Another well-established method employed to align features extracted from two domains consists of using a domain-adversarial training approach as first introduced by \cite{Ganin2015DomainAdversarialTO}. It works by adding a new, small classifier on top of the existing model which serves as a domain discriminator. The role of this classifier is to predict the original domain given the features extracted by the backbone, i.e. whether \(\mathcal{F}({X}_{t})\) and \(\mathcal{F}({X}_{aug})\) come from $D_{t} = (\mathcal{X}_{t}$) or $D_{aug} = (\mathcal{X}_{aug}$).

The domain discriminator $\mathcal{D}$ is trained using a confusion loss. Indeed, the objective is to confuse the detection model about the domain by maximizing the domain discriminator loss while minimizing the original detection loss. The student model tries to fool the domain discriminator by generating features that are more difficult to identify whether an image comes from $D_{t}$ or $D_{aug}$, playing a min-max adversarial game. As the gradient of the domain discriminator loss would typically prompt the model to improve at distinguishing between the domains, a gradient reversal layer (GRL) \cite{ganin2015unsupervised} is incorporated in order to invert the sign of the gradient during backpropagation through the domain classifier. This effectively makes the gradient of the domain classifier loss negative, leading the model to attempt to maximize the domain classifier loss rather than minimize it. This encourages the model to learn features that are invariant to the domain.

The loss associated with the domain discriminator is obtained by a binary cross entropy loss as given by:   
\begin{equation}
    \label{eq:adv_loss}
    \mathcal{L}_{\text{adv}} = - \sum_{i}^N d_i \cdot \log(\mathcal{D(\mathcal{F}}(x_i)))\ + (1- d_i) \cdot \log(1 - \mathcal{D}(\mathcal{F}(x_i))),
\end{equation} 
where $i$ indicates the $i^{th}$ image in the batch of size $N$, composed of both images sampled from $\mathcal{X}_{t}$ and $\mathcal{X}_{aug}$ images, and $d_i$ represents the domain label of $x_i$ (either 0 for $D_{t}$ or 1 for $D_{aug}$). 

\begin{figure}
	\centering
    \includegraphics[width=0.97\textwidth]{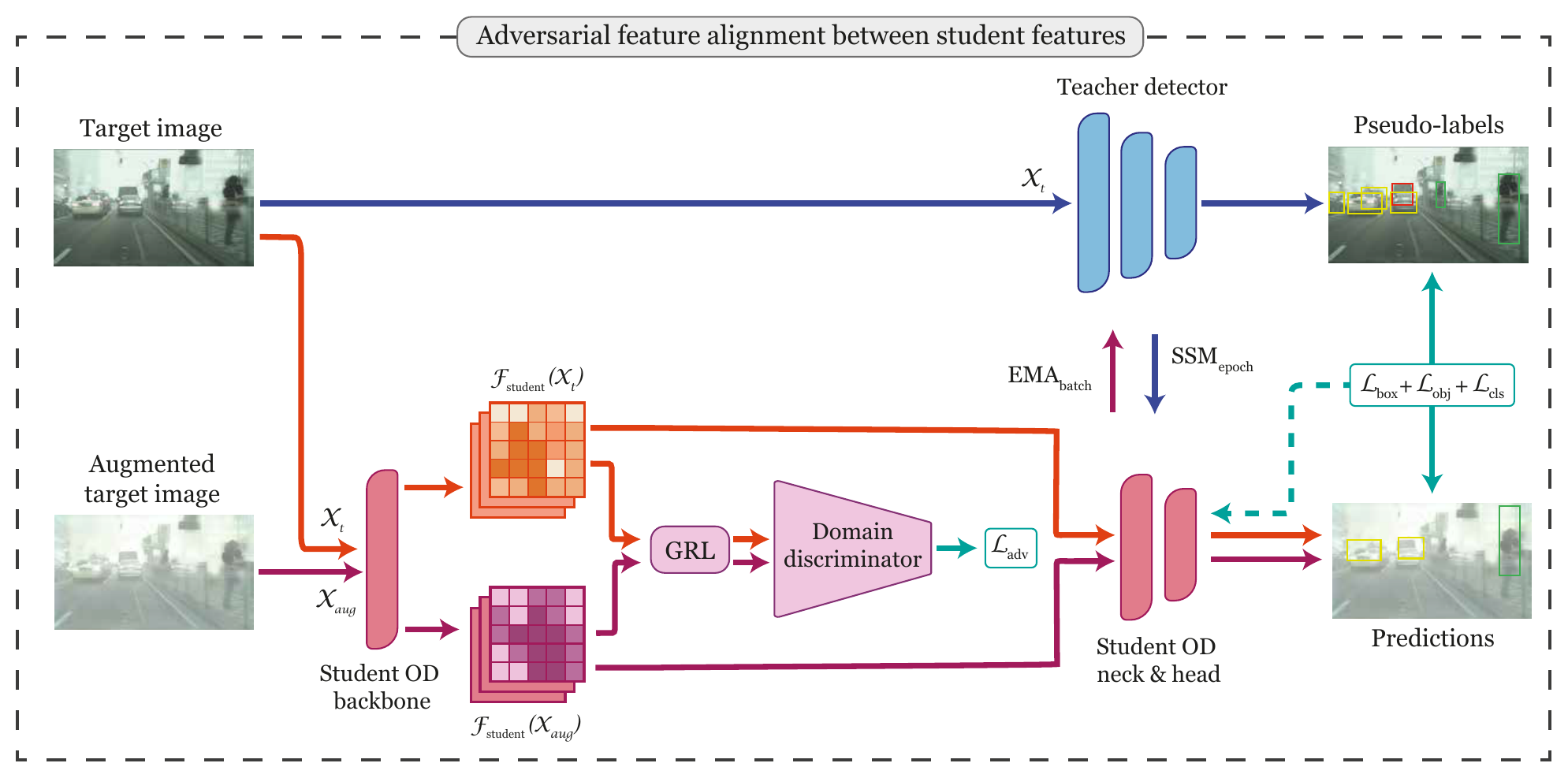}
    \parbox{1\textwidth}{\caption{Adversarial-based feature alignment between the student features. For this setting, the original SF-YOLO framework is modified so the student OD model receives both the target images and the augmented ones produced by the Target Augmentation Module (which is not represented in this figure) to allow the alignment process at the student model's level.
  }\label{fig:ADV_align_student}}
\end{figure}

Finally, the objective loss for the SF-YOLO method using an adversarial feature alignment strategy is given by:
\begin{equation}
    \max_{\mathcal{D}}\min_{\phi} \lambda_{\text{b}} \mathcal{L}_{\text{box}} + \lambda_{\text{c}} \mathcal{L}_{\text{cls}} + \lambda_{\text{d}} \mathcal{L}_{\text{obj}} - \lambda_a \mathcal{L}_{\text{adv}}, \
    \label{eq:yolo_loss_adv}
\end{equation}
where $\mathcal{L}_{\text{box}}$, $\mathcal{L}_{\text{cls}}$, and $\mathcal{L}_{\text{obj}}$ represent the bounding box regression, classification, and objectiveness losses, respectively, while $\mathcal{L}_{\text{adv}}$ denotes the adversarial loss. The $\lambda$ terms are weighting hyperparameters that control the relative importance of each loss component in the overall objective function. $\lambda_a$ serves as a trade-off parameter to balance the adversarial training, and $\phi$ denotes the whole student detection model composed of the backbone $\mathcal{F}$, the neck, and the head as described in Figure \ref{fig:ADV_align_student}. The negative sign in front of $\mathcal{L}_{\text{adv}}$ is due to adversarial training.

\subsection{Results and discussion}

\begin{table}[!h]
	\centering
    \caption{Impact of Feature-Alignment Strategies on Detection Accuracy. The mAP for the C2F scenario and the car AP for the K2C and S2C scenarios are reported. The "Source only" method refers to the source model without adaptation. AA stands for Adversarial Alignment and GWA for GW Alignment.}
	\label{tab:feat-align}
	\resizebox{\textwidth}{!}{%
	\begin{tabular}{
		c
        @{\hspace{4pt}} | @{\hspace{4pt}}
		c
        @{\hspace{4pt}} | @{\hspace{4pt}}
		c
        @{\hspace{4pt}} | @{\hspace{4pt}} 
		c
		c
		c
	}
	\toprule
	\textbf{Features aligned} & \textbf{Method} & \textbf{Detector} & \textbf{C2F} & \textbf{K2C} & \textbf{S2C} \\
	\midrule
    - & Source only &  & 41.8 & 59.8 & 63.9  \\ 
    - & EMA &  & 44.4 & 62.2 & 69.7 \\ 
    - & EMA + SSM &  & 51.2 & 62.7 & 69.3 \\ 
    \(\mathcal{F}_{\text{student}}(\mathcal{X}_{t})\) and \(\mathcal{F}_{\text{student}}(\mathcal{X}_{aug})\) & EMA + AA & YOLOv5l & 44.3 & 60.6 & 69.7\\
	\(\mathcal{F}_{\text{student}}(\mathcal{X}_{t})\) and \(\mathcal{F}_{\text{student}}(\mathcal{X}_{aug})\) & EMA + SSM + AA &  & 49.0 & 61.3 & 67.9 \\
	\(\mathcal{F}_{\text{student}}(\mathcal{X}_{t})\) and \(\mathcal{F}_{\text{student}}(\mathcal{X}_{aug})\) & EMA + SSM + GWA &  & 48.2 & 61.8 & 67.3 \\
	\(\mathcal{F}_{\text{teacher}}(\mathcal{X}_{t})\) and \(\mathcal{F}_{\text{student}}(\mathcal{X}_{aug})\) & EMA + SSM + GWA &  & 50.7 & 62.5 & 69.4 \\   
    \midrule
    - & Source only &  & 27.4 & 35.9 & 48.8 \\ 
    - & EMA &  & 41.1 & 49.3 & 57.1 \\ 
    - & EMA + SSM &  & 42.5 & 49.4 & 57.7 \\ 
	\(\mathcal{F}_{\text{student}}(\mathcal{X}_{t})\) and \(\mathcal{F}_{\text{student}}(\mathcal{X}_{aug})\) & EMA + AA& YOLOv5s & 32.8 & 46.6 & 57.3 \\
	\(\mathcal{F}_{\text{student}}(\mathcal{X}_{t})\) and \(\mathcal{F}_{\text{student}}(\mathcal{X}_{aug})\) & EMA + SSM + AA&  & 38.4 & 47.7 & 57.7 \\
	\(\mathcal{F}_{\text{student}}(\mathcal{X}_{t})\) and \(\mathcal{F}_{\text{student}}(\mathcal{X}_{aug})\) & EMA + SSM + GWA &  & 38.1 & 48.0 & 57.1 \\
	\(\mathcal{F}_{\text{teacher}}(\mathcal{X}_{t})\) and \(\mathcal{F}_{\text{student}}(\mathcal{X}_{aug})\) & EMA + SSM + GWA &  & 41.6 & 49.7 & 57.5 \\   
	\bottomrule
	\end{tabular}
	}
\end{table}

While aligning the features extracted from the source and target domains might help in UDA as shown by \cite{chen2018domain,Saito2018StrongWeakDA,Wang2021AFANAF}, we observe in Table \ref{tab:feat-align} that aligning target domain features with their augmented counterparts does not lead to an increase in performance in our SFDA case but rather preserves or even slightly decreased it. Note that aligning the features extracted by the teacher and the student \(\mathcal{F}_{\text{teacher}}(\mathcal{X}_{t})\) and \(\mathcal{F}_{\text{student}}(\mathcal{X}_{aug})\) gives better performance compared to the direct alignment of the features extracted by the student \(\mathcal{F}_{\text{student}}(\mathcal{X}_{t})\) and \(\mathcal{F}_{\text{student}}(\mathcal{X}_{aug})\). We observed both of those trends on the three different SFDA scenarios and for the two model sizes, the large and small YOLOv5 models.

To conclude, our experiments in Table \ref{tab:feat-align} suggest that explicit feature alignment between the domains $D_{t}$ and $D_{aug}$ is too aggressive for the SFDA setting. Instead, the combination of EMA and SSM implicitly achieves a weaker form of alignment that yields the best performance. This contrasts with what is usually observed in UDA, where aligning source and target feature distributions has proven beneficial \cite{chen2018domain,Saito2018StrongWeakDA,Wang2021AFANAF}.
We hypothesize that, as in the SFDA paradigm, the pre-trained source model provides the only connection to the source domain, aligning the feature spaces of augmented target views, which may deviate too much from the source distribution, risks degrading the source model's performance rather than improving it. The source model specializes in the source domain, so forcing alignment to different feature distributions appears to cause more harm than good in this setting. As a result, we opted against using feature alignment to keep our method simpler.

\bibliographystyle{splncs04}
\bibliography{egbib}
\end{document}